\documentclass[journal]{IEEEtran}
\usepackage{amsmath, amsfonts}

\usepackage{algorithm}
\usepackage{array}
\usepackage[caption=false, font=footnotesize, labelfont=rm, textfont=rm]{subfig}
\usepackage{textcomp}
\usepackage{stfloats}
\usepackage{url}
\usepackage{verbatim}
\usepackage{graphicx}
\usepackage{cite}
\usepackage[table]{xcolor}
\usepackage{xcolor}

\usepackage{xcolor}
\usepackage{booktabs}
\usepackage{ragged2e}
\usepackage{booktabs, makecell, multirow, tabularx}
\usepackage{siunitx}
\usepackage{setspace}
\usepackage{varwidth}
\usepackage{algorithmicx}
\usepackage{algpseudocode}
\usepackage{algorithmicx, algorithm}
\usepackage{float} 
\usepackage{adjustbox}
\usepackage{bm}
\usepackage{balance}
\usepackage{xspace}

\usepackage[
    colorlinks, 
    linkcolor=black,
    anchorcolor=black,
    citecolor=black,
    urlcolor=black
]{hyperref}

\makeatletter
\algrenewcommand\alglinenumber[1]{\textbf{#1.}} 
\makeatother

\newcommand{\srcfea}{f^{S}} %
\newcommand{\tgtfea}{f^{T}} %
\newcommand{\backbone}{\mathcal{F}} %
\newcommand{\dethead}{\mathcal{F}_{D}} %
\newcommand{\evihead}{\mathcal{F}_{E}} %
\newcommand{\loss}{\mathcal{L}} 
\newcommand{\IMGDisc}{\mathcal{D}_{\textit{IMG}}} 
\newcommand{\Disc}{\mathcal{D}} 
\newcommand{\src}{\mathcal{X}^S}
\newcommand{\tgt}{\mathcal{X}^T}

\newcommand{\lambdaSHFA}{{\lambda}_{\textit{SHFA}}}
\newcommand{\lambdaRSAA}{{\lambda}_{\textit{RSAA}}}

\newcommand{\lambdaSecmask}{{\lambda}_{\textit{se}}}

\newcommand{\lossEVI}{\mathcal{L}_{\textit{EVI}}}
\newcommand{\lossSHFA}{\mathcal{L}_{\textit{SHFA}}}
\newcommand{\lossRSAA}{\mathcal{L}_{\textit{RSAA}}}
\newcommand{\lossINS}{\mathcal{L}_{\textit{INS}}}
\newcommand{\lossIMG}{\mathcal{L}_{\textit{IMG}}}
\newcommand{\lossSRC}{\mathcal{L}_{\textit{S}}}

\def\ourmethod{{\textit{CR-Net}}\xspace}
\def\ourda{{\textit{SHFA}}\xspace}
\def\ouradja{{\textit{RSAA}}\xspace}

\def\ourinsfactor{{\text{reliable instance factor}}\xspace}
\def\ouradjafactor{{\text{secure adjacency factor}}\xspace}

\begin{document}
    %
    \title{Cross-Resolution SAR Target Detection Using Structural Hierarchy Adaptation and Reliable Adjacency Alignment}
    %
    %
    %

    \author{Jiang~Qin,~\IEEEmembership{Student~Member,~IEEE,~}
    Bin~Zou,~\IEEEmembership{Senior~Member,~IEEE,~}
    \\Haolin~Li,~\IEEEmembership{Student~Member,~IEEE,~} Lamei~Zhang,~\IEEEmembership{Senior~Member,~IEEE,~}

    \thanks{This work is partly supported by the National Natural Science
    Foundation of China under Grant 62271172. \textit{(Corresponding author: Bin~Zou)}}

    \thanks{All authors are with the Department of Information Engineering, Harbin
    Institute of Technology, Harbin 150001, China (e-mail: qinjiang19b@163.com;
    zoubin@hit.edu.cn; lhl\textunderscore hit@163.com; lmzhang@hit.edu.cn)}

    }

    \markboth{Journal of \LaTeX\ Class Files,~Vol.~14, No.~8, August~2015}%
    {Shell \MakeLowercase{\textit{et al.}}: Bare Demo of IEEEtran.cls for IEEE Journals}

    \maketitle

    \begin{abstract}
        In recent years, continuous improvements in SAR resolution have significantly benefited applications such as urban monitoring and target detection. However, these improvements in resolution have also led to increased discrepancies in scattering characteristics, posing challenges to the generalization ability of target detection models.
        While domain adaptation technologies provide a potential solution, the inevitable discrepancies caused by resolution differences often result in blind feature adaptation and unreliable semantic propagation, ultimately degrading the domain adaptation performance.
        To address these challenges, this paper proposes a novel SAR target detection method, termed \ourmethod, which incorporates structure priors and evidential learning theory into the detection model, enabling reliable domain adaptation for cross-resolution detection.
        To be specific, \ourmethod integrates Structure-induced Hierarchical Feature Adaptation (\ourda) and Reliable Structural Adjacency Alignment (\ouradja).
        The \ourda module is designed to establish structural correlations between targets and achieve
        structure-aware feature adaptation, thereby enhancing the interpretability of the adaptation process.
        Afterwards, the \ouradja module is proposed to enhance reliable semantic alignment, by leveraging the secure adjacency set to transfer valuable discriminative knowledge from the source domain to the target domain. This further improves the discriminability of the detection model in the target domain.
        Based on experimental results from different-resolution datasets, the proposed \ourmethod significantly enhances cross-resolution adaptation by preserving intra-domain structures and improving discriminability. It achieves state-of-the-art (SOTA) performance in cross-resolution SAR target detection.

    \end{abstract}


    \IEEEpeerreviewmaketitle

    \section{Introduction}
    \begin{figure}[!t]
        \centering
        \includegraphics[width=0.999\columnwidth]{
            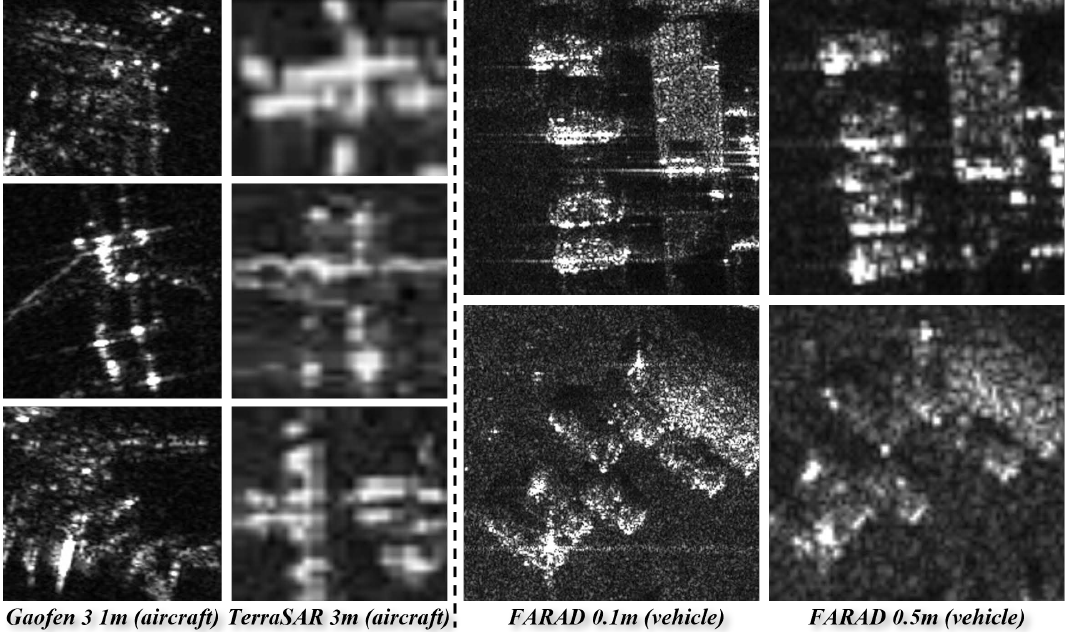
        }
        \vspace{-7mm}
        
        \caption{Illustrating the significant scattering and scale
        differences in different-resolution SAR images. Targets in high-resolution
        images typically display finer scattering characteristics and more complex
        structural details, with clearer textures, and multiple scattering points.
        In contrast, targets in low-resolution images experience a substantial loss
        of detailed information and are often represented by blurred contours or
        key scattering points. }
        \label{fig:introduction}
        \vspace{-5mm}
    \end{figure}

    \IEEEPARstart{S}{ynthetic} Aperture Radar (SAR) achieves high-resolution
    imaging by actively transmitting and receiving electromagnetic waves. 
    This unique capability allows SAR to operate effectively under adverse weather conditions
    and regardless of daylight, enabling all-weather and all-day imaging.
    These attributes have made SAR widely applicable in the field of Earth observation~\cite{sartutorials,Liu04072025}.
    As one of the primary means of Earth observation, SAR target detection has been
    a hot topic in both research and applications. 
    In recent years, advancements in SAR sensor technology have significantly improved the spatial resolution of SAR images, transitioning from traditional meter-level resolution to sub-meter resolution. 
    This leap forward has unlocked significant potential for more refined
    Earth observation applications: sub-meter resolution SAR images can capture
    more target details, offering substantial advantages in fine-grained tasks
    such as urban monitoring~\cite{sartutorials}, disaster assessment~\cite{zhang2023frequency}, and target recognition~\cite{AIRcraft-1.0}.

    However, the improvement in resolution also introduces new challenges, particularly regarding the model generalization problem in cross-resolution target detection tasks. 
    As illustrated in Fig.~\ref{fig:introduction}, differences in resolution cause the same SAR target to exhibit significant variations in scattering characteristics and scales.
    For instance, targets in high-resolution images typically display finer scattering  characteristics and more complex structural details, with clearer relationships
    among edges, textures, and multiple scattering points.
    In contrast, targets in low-resolution images suffer from substantial detail loss and are generally characterized by blurred contours or prominent scattering centers.
    Restricted by these substantial discrepancies, target detection models trained on a single resolution struggle to extract discriminative features from other resolution SAR images.
    Specifically, when trained on low-resolution data, models tend to rely on the relatively limited structural information, resulting in the extracted features that often fail to accurately describe the complex details present in high-resolution SAR targets.
    Similarly, models trained on high-resolution SAR data also perform poorly on low-resolution SAR images, as their reliance on the abundant features in high-resolution SAR images leaves them lacking sufficient discriminative ability when handling low-resolution SAR targets.
    This bidirectional decline in generalization performance poses significant challenges for cross-resolution SAR target detection.

    These challenges become particularly evident during the resolution upgrades of SAR systems. 
    As SAR sensors advance from low to high resolution, existing detection models often struggle to adapt to the newly acquired high-resolution data, making it difficult to enable a smooth transition in system deployment and applications. 
    An alternative approach involves manually annotating datasets to train detection models for different resolutions. 
    However, this process, along with creating separate annotation for varying resolution datasets, is highly resource-intensive and significantly limits the feasibility of joint analysis SAR data with different resolutions. 
    Consequently, achieving effective generalization in cross-resolution target detection models and addressing the domain discrepancies  introduced by resolution differences have become critical challenges in the field of SAR target detection.

    Domain adaptation is an effective approach to address the aforementioned domain discrepancies and improve the generalization performance on the unseen data~\cite{DAN,DANN}.
    The goal of this technique is to transfer the knowledge learned from the labeled source domain to an unlabeled target domain with distribution differences, by enhancing the model's transferability and discriminability to gain satisfactory performance on the target domain data.
    
    In recent years, some advanced researches have leveraged domain adaptation techniques in SAR target detection, focusing on improving cross-modality (e.g. optical-to-SAR~\cite{ida,HSA,huang2024domain} and SAR-to-SAR~\cite{zhang2025cross,shi2021unsupervised,zou2023cross}) detection performance. 
    {For instance, Zhao~et~al.~\cite{zhao2022automatic,zhao2022feature} achieved SAR-to-SAR target detection by employing adversarial learning and transferable feature decomposition to extract domain-invariant features. However, such blind adaptation methods do not explicitly consider intra-domain variations, which may lead to feature misalignment and structural distortions within domains. 
    To address this issue, Pan~et~al.\cite{ida} proposed an unbalanced adversarial feature alignment approach, enabling ship detection from optical to SAR images. Nevertheless, some studies\cite{HTCN,kim2019self} have indicated that adversarial learning-based domain adaptation methods negatively affect feature discriminability. To mitigate this problem, several works~\cite{kim2019self,shi2021unsupervised,zou2023cross} have incorporated pseudo-label learning to enhance feature discriminability. However, unreliable predictions in the target domain may lead to risky semantic propagation and error accumulation during the pseudo-label learning process. 
    To reduce domain discrepancies between different sensors, other studies~\cite{zhang2025cross,zhang2023frequency,wang2018sar} have leveraged the physical priors of SAR targets to enhance transferability. Wang~et~al.\cite{wang2018sar} utilized sub-aperture decomposition\cite{SCEDet} and data augmentation to achieve knowledge transfer from synthetic data to real-world data. Some researchers~\cite{zhang2025cross} have introduced key scattering points of targets to enable center-aware feature adaptation, while Zhang~et~al.~\cite{zhang2023frequency} incorporated Fourier transform features~\cite{10934049} to establish local structural relationships and enable target detection from optical to SAR images. However, the feasibility of these methods for cross-resolution detection in SAR images has not yet been verified. }

    Although the aforementioned methods have made some progress in cross-domain SAR target detection, applying these black-box style models to cross-resolution SAR target detection still faces several critical challenges:

    \begin{itemize}
        \item[\textit{1)}] \textit{Structural Distortions by the Blind Adaptation:} 
            The intra-domain SAR targets often exhibit significant structural diversity (illustrated in Fig~\ref{fig:motivation}) due to variations in imaging angles, target categories, and scattering characteristics.
            Existing domain adaptation methods typically aim to minimize domain gaps by unconditionally aligning the feature distributions between source and target domains.
            However, this structure-agnostic and blind distribution alignment lack interpretability and overlook dataset-specific structural variations, potentially distorting critical distribution structures. 
            These structural distortions degrade the discriminability of target domain features for detection tasks. 
            To address this, it is essential to introduce structure-aware methods that can quantify and leverage these structural differences, guiding the domain adaptation process to
            alleviate distribution distortions and preserve the discriminability.
    \end{itemize}

    \begin{itemize}
        \item[\textit{2)}] \textit{Risky Semantic Propagation by the Unreliable Predictions:} In unsupervised domain adaptation, the model relies heavily on the source domain for discriminating targets, often resulting in unreliable  predictions for the unlabeled target domain. These unreliable predictions may serve as erroneous reference points during the adaptation  process, propagating the semantic information to incorrect target instances. 
        Such risky semantic propagation further amplifies prediction unreliability and degrades the discriminability, ultimately leading to suboptimal performance in the target domain. 
        Consequently, improving the reliability of the adaptation process and promoting reliable semantic discriminability are essential for mitigating error propagation and achieving robust domain adaptation.
    \end{itemize}

    To address the aforementioned challenges, this paper proposes \ourmethod for cross-resolution SAR target detection that incorporates evidential learning theory and structural constraints. This approach effectively mitigates the issues of structural distortion and risky semantic propagation caused by traditional blind feature adaptation. The main contributions of this paper are as follows:
    \begin{itemize}
        \item[1)] We incorporate evidence learning theory into the SAR target detection model by leveraging the Dirichlet distribution to estimate uncertainty. This enables the model to perceive uncertainty in its predictions, laying a solid foundation for subsequent reliable domain adaptation.
    \end{itemize}
    \begin{itemize}
        \item[2)] A Scattering Structure Distance (SSD) is proposed to quantify the structure similarity between SAR targets. Leveraging the scattering structure similarity of instances, a Structure-induced Hierarchical Feature Adaptation (\ourda) module is developed. By introducing structural constraints, \ourda module facilitates hierarchical feature adaptation between the source and target instances,  enhancing  the feature transferability and avoiding structural distortions throughout the adaptation process.
    \end{itemize}
    \begin{itemize}
        \item[3)] A Reliable Structural Adjacency Alignment (\ouradja) module is proposed to capture reliable adjacency relationships by aligning semantic information between reliable neighbors across domains. \ouradja module effectively transfers discriminative semantics from the source domain to the target domain, significantly improving the model's performance on the target domain while mitigating error accumulation.
    \end{itemize}
    \begin{itemize}
        \item[4)] The proposed \ourmethod combines \ourda and \ouradja modules, achieving significant performance improvements in cross-resolution SAR target detection tasks. 
        Experimental results demonstrate that \ourmethod not only preserves the intra-domain structures but also significantly enhances the reliability and accuracy of cross-resolution detection.
    \end{itemize}

    The remainder of this paper is organized as follows. Section \ref{related work} reviews related works. Section \ref{evidential learning} explains
    how evidence learning is incorporated into SAR detection models. Building on the evidence-enhanced SAR detection framework, Section \ref{Methodology}
    details the proposed cross-resolution detection method. Section \ref{Experiments} presents the experimental results and analysis. Finally, Section \ref{conclusion} concludes the paper.

    \section{Related Works}
    \label{related work}
    \subsection{Uncertainty Estimation and Evidential Learning}
    Uncertainty is a key factor for evaluating the 
    predictive reliability of deep models~\cite{sensoy2018evidential,pei2024evidential}. 
    Bayesian deep learning~\cite{maddox2019simple} and subsequent approximate methods~\cite{gal2016dropout,lakshminarayanan2017simple} provide theoretical and practical approaches for uncertainty modeling. However, they still face challenges related to high computational cost and low efficiency.
    Recently, evidence learning~\cite{sensoy2018evidential} has gained significant attention due to its capability to explicitly model and inference with evidence, making it a powerful approach for quantifying uncertainty. 
    Compared to traditional Bayesian methods, evidence learning learns a Dirichlet distribution to simultaneously output predictive confidence and uncertainty, avoiding high computational costs and improving  flexibility~\cite{zhang2023provable}. Currently, evidential learning has been widely applied to high-reliability tasks such as medical diagnosis~\cite{10656006} and anomaly detection~\cite{10948323,zhou2024outlier}. However, its application in cross-resolution SAR target detection remains largely unexplored. This paper applies evidence learning in cross-resolution SAR detection tasks to model uncertainty and improve adaptation reliability across resolutions.

    \subsection{SAR Target Detection}
    In SAR target detection, mainstream approaches are typically divided into one-stage detectors (e.g., RetinaNet~\cite{focal}, YOLOs~\cite{yolov7}) and two-stage detectors (e.g., Faster RCNN~\cite{fasterrcnn}, Sparse RCNN~\cite{sparsercnn}). 
    {Building on these frameworks, the SAR community has proposed various advancements. For example, Zhou~et~al.~\cite{10091564} developed a sidelobe-aware network for better small ship detection, while Ju~et~al.~\cite{10285445} and Zhao~et~al.~\cite{10056331} improved rotated target detection through polar encoding and enhanced inter-class separability~\cite{qin2024scattering}. 
    Yang~et~al.~\cite{10813601} proposed a dynamic detection framework to boost dense prediction performance in complex scenarios. 
    For two-stage detectors, Cui~et al.~\cite{8763918} incorporated attention mechanisms~\cite{10294268} for improved multi-scale detection, and Cheng~et~al.~\cite{cheng2022inshore} addressed nearshore detection with a saliency enhancement algorithm.
    Some studies~\cite{ren2025confucius,9916304} have leveraged generative adversarial networks~\cite{goodfellow2014generative} (GANs) to enhance model discrimination and robustness. For instance, Ren~et~al.~\cite{ren2025confucius} introduced a Confucius tri-learning framework, which jointly trains two classifiers and a generator to achieve better discrimination performance with limited samples. Ju~et~al.~\cite{9916304} employed GANs to suppress background interference, thus improving detection in complex environments. Some researchers~\cite{8950292} utilized GANs to guide models in extracting essential structural features, thereby boosting discriminative capability.
    Other works~\cite{10054495,10770564,SCEDet} have focused on improving detection performance by mitigating environmental clutter~\cite{10054495}, multi-scale dynamic learning~\cite{10770564}, and optimizing data quality with data transformations or sub-aperture decomposition~\cite{10623223,SCEDet}.}

    {Although these detection methods perform well when the training and testing data are independently and identically distributed, their performance drops significantly when confronted with data from different sensors, modalities, or resolutions.}

    \subsection{Unsupervised Domain Adaptation}
    Unsupervised Domain Adaptation (UDA) is an important research direction in transfer learning, aiming to improve the performance of models on the target domain by leveraging labeled data from the source domain and unlabeled data from the target domain~\cite{DAN}. The core challenge lies in the distribution discrepancy (i.e., domain shift) between the source and target domains. Researchers have proposed various methods to reduce this discrepancy, including distribution alignment (e.g., DAN~\cite{DAN} and MADA~\cite{MADA}), pseudo-label generation~\cite{khodabandeh2019robust,kim2019self},
    image translation methods~\cite{HTCN,qin2024conditional},
    and mean-teacher training methods~\cite{deng2021unbiased, cat}. 
    UDA is widely applied in fields such as computer vision~\cite{deng2023harmonious,lu2025visual,sadis}, medical image analysis~\cite{8995481}, and autonomous driving~\cite{cat}. 
    Despite significant progress, UDA still faces challenges such as    interpretability~\cite{yue2021transporting,wang2025crowdvlm}, domain structure distortion~\cite{DAN},
    and unreliable knowledge transfer~\cite{pei2024evidential}.

    \subsection{UDA for SAR Target Detection}
    In recent years, UDA has become a popular research topic in SAR target detection. Mainstream methods include image translation~\cite{HTCN,10767752}, adversarial adaptation~\cite{shi2022unsupervised,10570480}, and pseudo-label learning~\cite{zou2023cross,shi2021unsupervised}. 
    {Image translation-based UDA~\cite{10767752} is a two-stage approach. Researchers have adopted CycleGAN~\cite{cyclegan,shi2022unsupervised,shi2021unsupervised} to generate translated images that reduce domain discrepancies, followed by domain adaptation techniques to enhance detection performance. Zhang~et~al.~\cite{HSA} created pseudo-SAR images using the Fourier transform for feature alignment, while others~\cite{huang2024domain} added speckle noise to optical images for further mitigate domain discrepancies. To improve detection on low-quality SAR images, Pu~et~al.~\cite{pu2023ship} employed CycleGAN for image enhancement, thereby boosting target detection accuracy. However, as these two-stage approaches are typically not trained in an end-to-end manner, it is difficult to guarantee optimal performance.
    In contrast, adversarial adaptation aligns domain feature distributions via end-to-end adversarial learning~\cite{DANN}. Some efforts integrate adversarial adaptation with feature decoupling~\cite{zhao2022feature} or context relationships~\cite{zhao2022automatic} to facilitate knowledge transfer. Liu~et~al.~\cite{10570480} leverage semantic calibration and adversarial feature alignment to enhance cross-domain ship detection. However, some studies have revealed that such blind adversarial training may lead to distortion of domain structures and impair the discriminative capability of learned features.
    To alleviate the degradation of feature discriminability caused by adversarial learning, some works~\cite{zou2023cross, shi2021unsupervised, shi2022unsupervised} have proposed incorporating pseudo-label learning strategies into the domain adaptation process.
    However, these methods essentially rely on the discriminative capability learned from the source domain data to select pseudo-labels of the target domain data. As a result, the discriminative capability of such UDA methods remains overly biased toward the source domain.  Directly applying such biased knowledge to the target domain can result in unreliable predictions, and aligning these unreliable predictions may even cause risky semantic propagation and error accumulation.}

    \section{{Preliminaries}}
    \label{evidential learning}
    \begin{figure}[!t]
        \centering
        \includegraphics[width=0.9999\columnwidth]{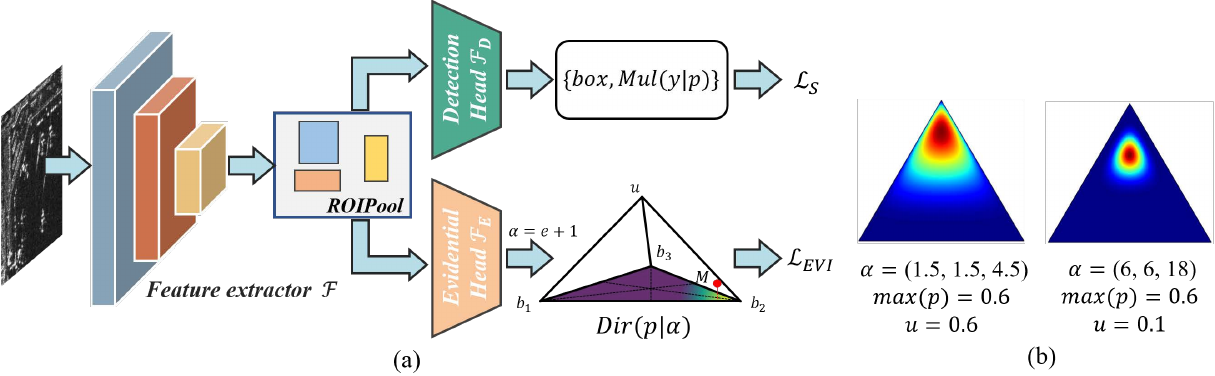}
        \vspace{-5mm}
        
        \caption{The architecture of the detection model enhanced by evidential learning. (a) illustrate the model design. (b) shows the probability simplex with the same probability $\max(p)=0.6$ but different uncertainties. The probability $\max(p)$ alone does not fully reflect the model’s confidence, while the uncertainty $u$ provides a more comprehensive assessment of the prediction reliability.}
        \label{fig:det}
        \vspace{-5mm}
    \end{figure}

    To facilitate reliable cross-domain knowledge transfer, we first integrate the evidential learning into the target detection model, as illustrated in Fig.~\ref{fig:det}. This model mainly consists of three parts: the
    feature extractor $\backbone$, the detection branch $\dethead$ and the evidential learning branch $\evihead$.

    \subsection{{Detection Model Enhanced by Evidential Learning}}
    \label{evidential modeling}
    The detection branch $\dethead$ is responsible for class prediction
    $y$ and $box$ regression. Since the target class prediction satisfies the characteristics
    of a Multinomial distribution, the class probability $p$ can be modeled as
    parameters of the Multinomial distribution~\cite{sensoy2018evidential,pei2024evidential}, which is expressed as: $y \sim P(y \mid p) = Mul(y \mid  p)$.

    The evidential learning branch $\evihead$, as shown in Fig.~\ref{fig:det}, is
    primarily designed to enhance the model's reliability and quantify uncertainty.
    By introducing the Dirichlet distribution, the model can represent class probabilities in a way that captures the confidence in its predictions. The Dirichlet distribution $Dir(p \mid \alpha)$, as a conjugate prior of the Multinomial distribution, is defined as:
    \begin{equation}
        \label{eq1}
        \begin{aligned}
            p & \sim P(p \mid \alpha) = Dir(p \mid \alpha) \\
              & = \begin{cases}\frac{1}{\beta(\alpha)}\prod_{c=1}^{C}p_{c}^{\alpha_c - 1},&\text{if }p \in S_{C}, \\ 0,&\text{otherwise}.\end{cases}
        \end{aligned}
    \end{equation}
    $\beta(\alpha)$ is the Beta function for the $C$-dimensional Dirichlet distribution.
    $\alpha=[\alpha_{1},...,\alpha_{C}]$ represents the concentration parameters of the Dirichlet distribution. $S_{C}$ is $\{C-1\}$-dimensional unit simplex.
    In evidential learning, the concentration parameter $\alpha=e+1$, where the the evidence score $e$ is predicted by the evidential learning branch $\evihead$. $e \geq 0$, representing the evidence for class predictions. The belief mass $b_{c}$
    and uncertainty $u$ for a sample $x$ are defined as:
    \begin{equation}
        \label{eq2}b_{c}= \frac{e_{c}}{S}, u=\frac{C}{S},
    \end{equation}
    where $S = \sum\nolimits_{c
    = 1}^{C}{{e_c}}+ C$, representing the Dirichlet strength. When the belief mass $b_{c}$ is larger, it indicates that
    more evidence $e_c$ is assigned to class $c$. The discriminative probability of $c\text{-th}$ class is $p_c=\alpha_c/S $.    
    A larger Dirichlet strength $S$ reduces the overall uncertainty $u$, increasing the confidence and reliability in class predictions.

    \subsection{Model Optimization Using Source Domain Data}
    Based on the Multinomial distribution $P\left(y \mid p\right) = Mul\left(y\mid p\right)$ and the Dirichlet distribution $P\left(p \mid \alpha\right) = Dir\left(p\mid\alpha\right)$, the process of predicting class label $y$ via Bayesian inference can be expressed as:
    \begin{equation}
        \label{eq3}P\left({y\mid \alpha }\right) = \frac{{P\left( {y,p\mid \alpha } \right)}}{{P\left( {p\mid \alpha ,y} \right)}}.
    \end{equation}
    According to Jensen's inequality and derivations, we have:
\begin{equation}
    \label{eq5}
    \begin{aligned}
        \log P\left({y\mid \alpha }\right)
        &\geq {\mathbb{E}_{p \sim Q\left( {p\mid \alpha } \right)} \log P\left({y\mid \alpha }\right)} \\
        &= \mathrm{KL}(Q(p\mid \alpha)\Vert P(p\mid \alpha,y)) \\
        &+ {\mathbb{E}_{p \sim Q\left( {p\mid \alpha } \right)}}\log P\left( {y\mid p} \right)  \\
        &+ {\mathbb{E}_{p \sim Q\left( {p\mid \alpha } \right)}}\log \frac{{P\left( {p\mid \alpha } \right)}}{{Q\left( {p\mid \alpha } \right)}}.
    \end{aligned}
\end{equation}
    Here, $p\sim Q(p\mid \alpha)$ is predicted by the evidential learning branch $\evihead$. 
    Due to the non-negativity of the KL divergence, the likelihood $\mathbb{E}_{p \sim Q(p\mid \alpha)} \log P(y\mid \alpha)$ can be maximized by optimizing the second and  third terms of Eq.~\ref{eq5}.
    
    The second term of Eq.~\ref{eq5} is the integral of the cross-entropy function over $Q(p\mid \alpha)$. By substituting the Dirichlet distribution $\mathrm{Dir}(p\mid \alpha)$ for $Q(p\mid \alpha)$, this term can be derived as
    \begin{equation}
        \label{eq6}
        \begin{aligned}
            {\mathbb{E}_{p \sim Q\left( {p\mid \alpha } \right)}}\log P\left({y\mid p}\right) = - \sum\limits_{c \in C}{{y_c}\left( {\left( {\psi (S) - \psi ({\alpha _c})} \right)} \right)}.
        \end{aligned}
    \end{equation}
    Here, digamma function $\psi \left( x \right) \approx \ln x - \frac{1}{{2x}}$.

    The third term of Eq.~\ref{eq5} represents an important prior regularization, which is equivalent to $\mathrm{KL}(Q(p\mid \alpha)\Vert P(p\mid \alpha))$. 
    Here, $P(p\mid \alpha)$ is the Dirichlet distribution, and its concentration parameter satisfies $\alpha \geq 1$.
    For class prediction problems, an ideal Dirichlet
    distribution $P(p\mid \alpha)$ should concentrate on the vertex of the simplex corresponding to the true class $y$. In other words, apart from class $y$, the concentration parameter $\alpha$ for other classes should be minimized to 1. Therefore, the prior regularization is formulated as:
    \begin{equation}
        \label{eq7}
        \begin{aligned}
            \mathrm{KL}(Q(p\mid \alpha)\ 
            &\Vert\ P(p\mid \alpha)) \\
            &\approx\ \mathrm{KL}\left({Dir}(p\mid \dot\alpha)\ 
            \Vert\ {Dir}(p\mid [1,\dots,1])\right)
        \end{aligned}
    \end{equation}

    Here $\dot \alpha ={y_c}+ \left({1 - {y_c}}\right){\alpha _c}$, where $\dot \alpha$ for the true label $y_{c}$ is set to 1,
    while $\dot \alpha$ for the remaining classes are
    maintained as predicted by the model. The goal is to minimize Eq.~\ref{eq7} to reduce evidence for incorrect classes.

    Given the source domain data $\cal{X}_{S}$, the overall optimization
    objective for the evidential learning branch is expressed as:
    \begin{equation}
        \label{eq8}
        \begin{aligned}
            {\lossEVI} & = \frac{1}{\left| \src \right|}\sum_{x \in \src}\sum_{c=1}^{C}y_{c}\left( \psi(S) - \psi(\alpha_{c}) \right) \\
                       & \quad + \mathrm{KL}\left( Dir(p \mid \alpha) \parallel Dir\left( p \mid [1, \ldots, 1] \right) \right).
        \end{aligned}
    \end{equation}

    For parameter optimization of the target task detection branch $\dethead$, the target detection loss $\lossSRC$ is directly used to optimize the model. For the evidential learning branch $\evihead$, its parameters are optimized by minimizing ${\lossEVI}$. Therefore, the total supervised loss for the source domain is presented as:
    \begin{equation}
        \label{eqsrc}
        \loss=\lossSRC+ {\lossEVI}.
    \end{equation}

    \section{Structural Hierarchy Adaptation and Reliable Adjacency Alignment}
    \label{Methodology}
    \begin{figure*}[!t]
        \centering
        \includegraphics[width=0.999\textwidth]{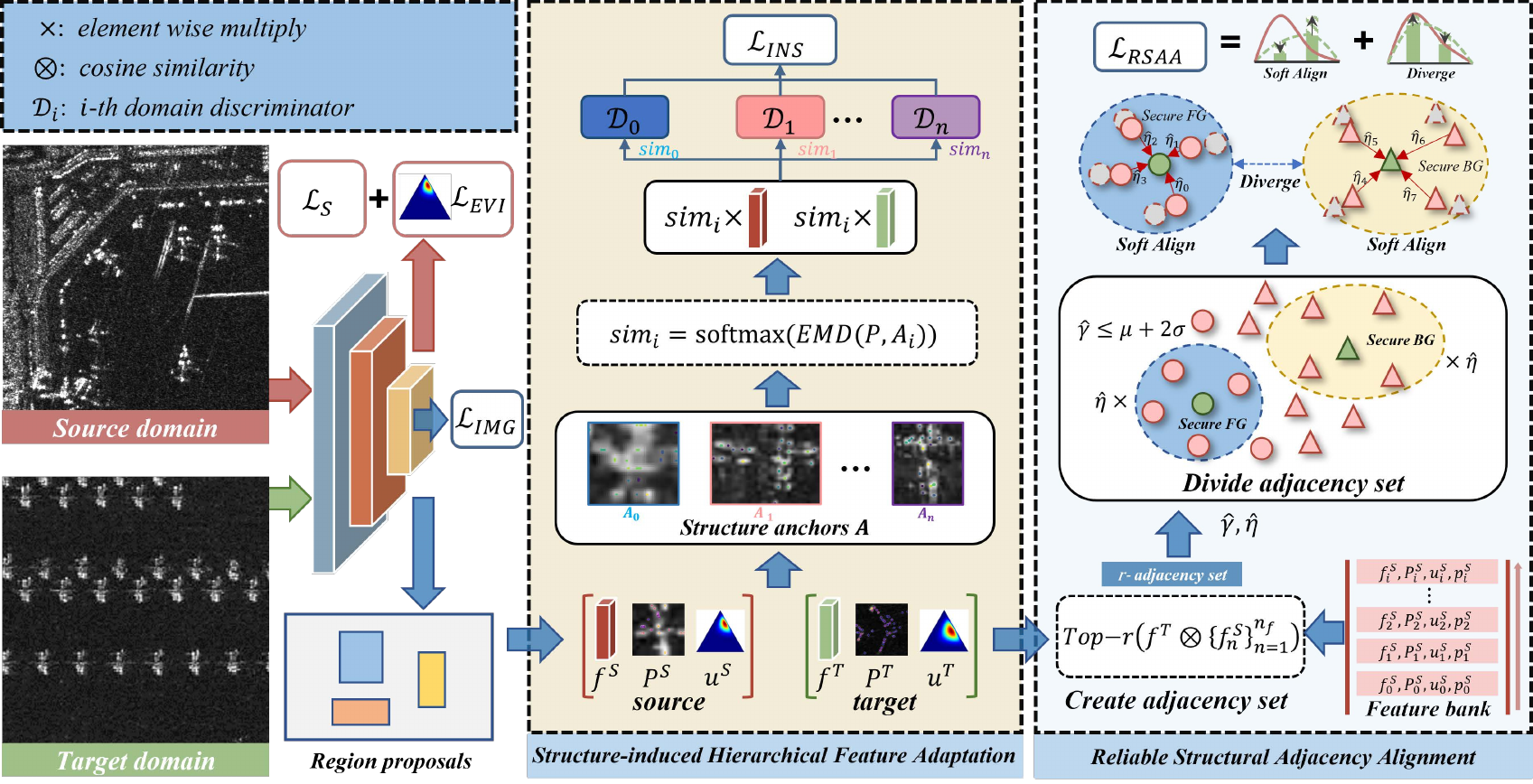}
        \caption{The overall framework of \ourmethod. Structure-induced Hierarchical Feature Adaptation (\ourda) leverages structural similarity to guide features from the source and target domains towards fine-grained feature distribution adaptation. 
        Reliable Structural Adjacency Alignment (\ouradja) utilizes structural similarity and uncertainty to select secure adjacency set, aligning the adjacency semantics to further enhance the detector's discriminability in the target domain.}
        \label{fig:frame}
        
    \end{figure*}

    \subsection{{Scattering Structure Distance}}
\begin{figure}[!t]
        \centering
        \includegraphics[width=0.9999\columnwidth]{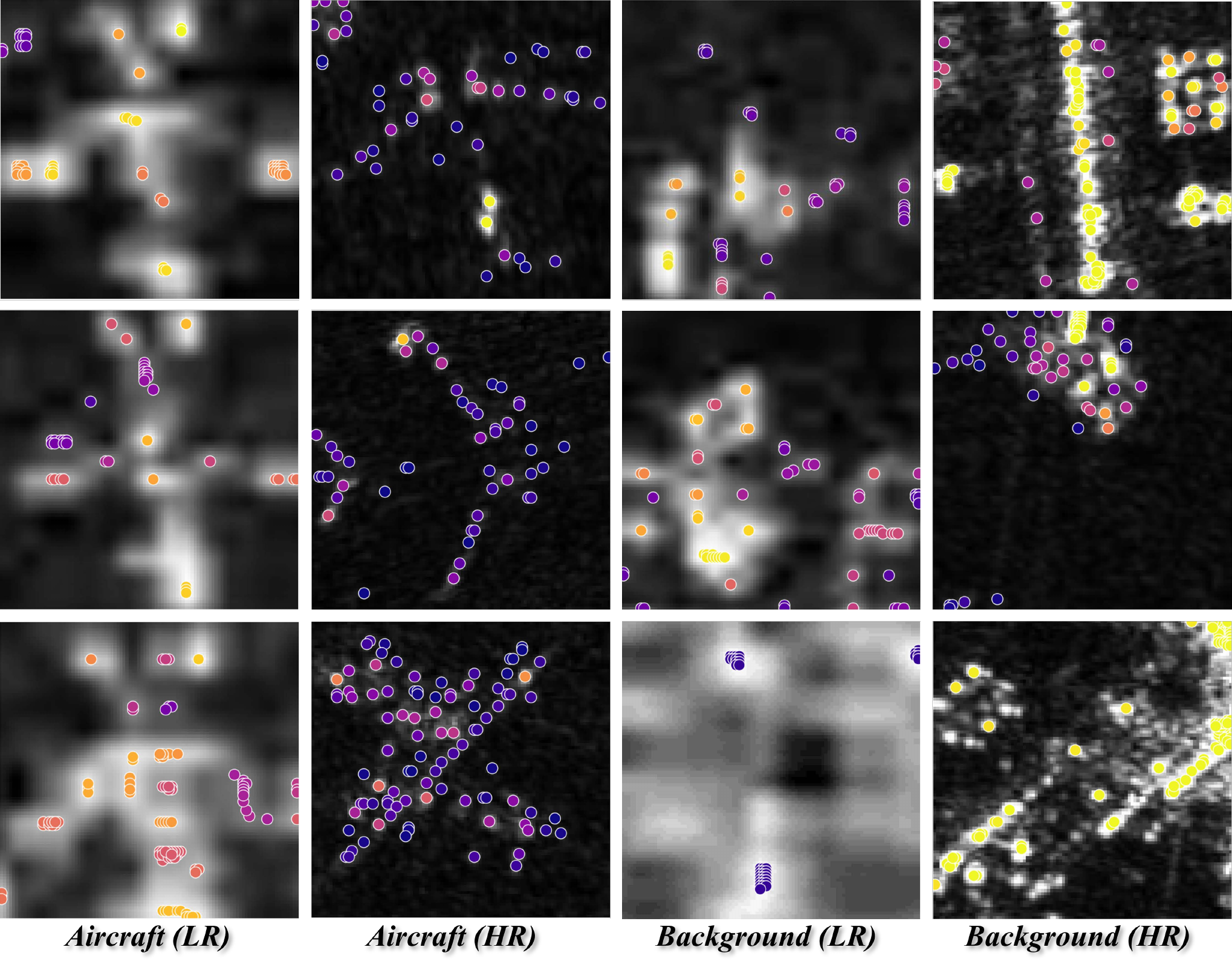}
        \vspace{-5mm}
        \caption{Scattering structures of foreground and background in low-resolution (LR) and high-resolution (HR) images. All images are resized to the same size for better comparison. Color indicates scattering intensity, ranging from blue (low) to yellow (high). Although the scattering features of LR and HR targets differ, the structural similarity remains(e.g. aircraft targets exhibit a cross-shaped pattern).}
        \label{fig:structures}
    \end{figure}

    SAR images are quite sensitive to natural structures, especially structures like dihedrals and trihedrals, which easily form strong scattering points due to their geometric stability. These points exhibit scale and resolution invariance~\cite{qin2024scattering,zhang2025cross}, resulting in topological structures that reveal correlations between targets across different resolutions. As shown in Fig.~\ref{fig:structures}, targets of the same category often display similar topological patterns (e.g., aircraft targets typically exhibit a cross-shaped structure). Quantifying these structural similarities can facilitate the association of targets at varying resolutions.
    
    Given a SAR target image $I_{i}$, the distribution of its strong scattering points is represented as:
    ${P_i}= \left\{{{x_n},{y_n},{a_n}}\right\}_{n = 1}^{N}$. $(x_{n},y_{n})$ and $a_{n}$ respectively represent the spatial coordinates and intensities of scattering points in the image. 
    Similarly, the scattering point set of another SAR target $I_{j}$ can be expressed as ${Q_j}= \left\{{{x_m},{y_m},{a_m}}\right\}_{m = 1}^{M}$.

    Considering the sparsity and uneven distribution of scattering points, traditional distance measurement methods cannot effectively quantify such distribution differences.
    In contrast, Earth Mover's Distance (EMD)~\cite{emd}, based on optimal transport theory~\cite{sinkhorn,feng2014note}, provides a more effective way to measure the distribution discrepancy between scattering point sets.
    {Specifically, EMD defines the distance between point sets $P_{i}$ and $Q_{j}$ as the minimum transportation cost required to transform the distribution $P_{i}$ into $Q_{j}$.
    It is robust to variations in the size and distribution, and it also can effectively capture both global and local differences between sparse scattering point sets under different resolutions and scales.
    Therefore, we propose to use EMD for calculating the scattering structure distance (SSD) between point sets.}
    
\begin{figure}[!t]
        \centering
        \includegraphics[width=0.9\columnwidth]{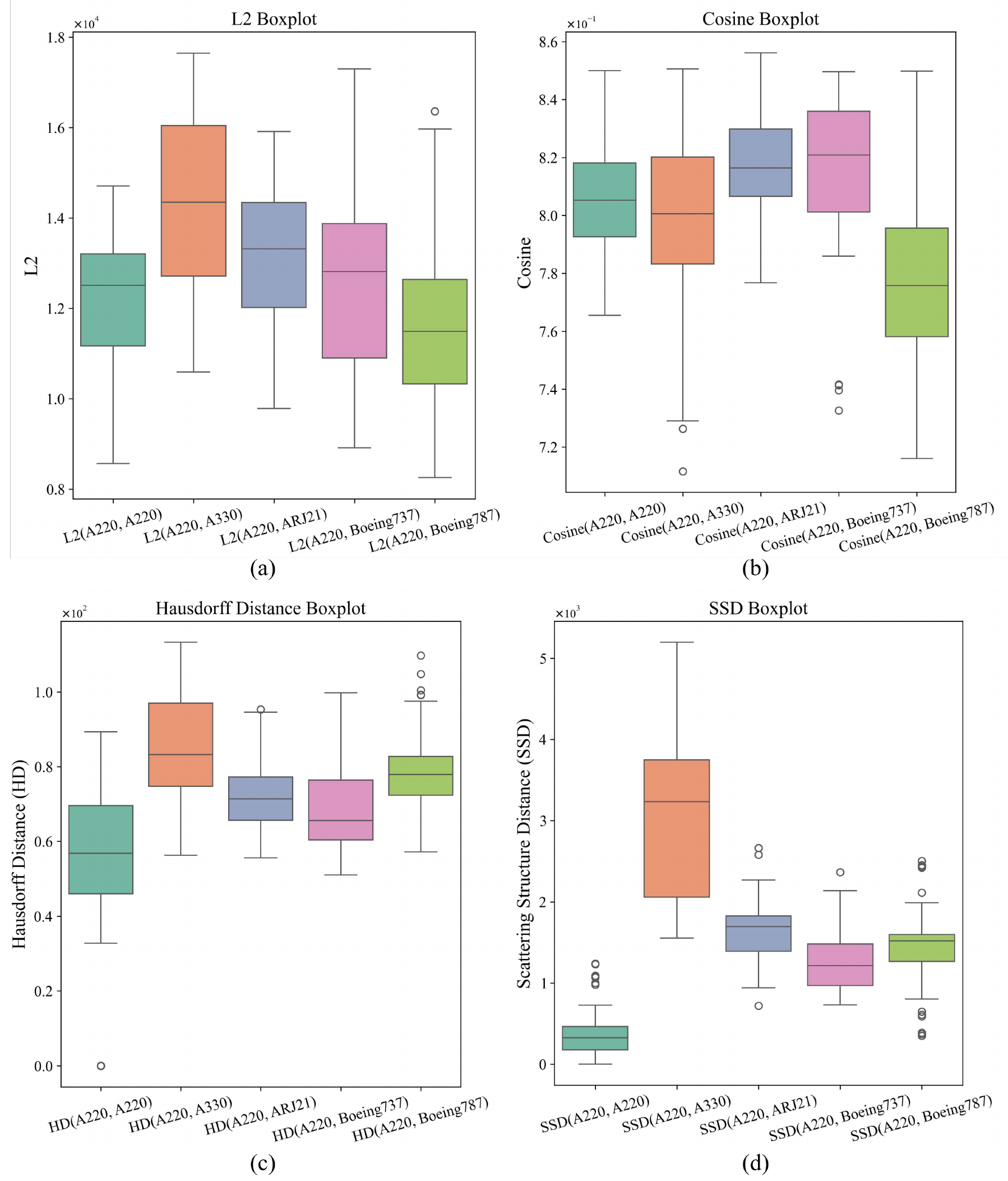}
        \caption{{Boxplots of different distance metrics for SAR target structures. (a)-(d) correspond to L2 distance, cosine distance, Hausdorff distance, and the proposed scattering structure distance (SSD), respectively. Compared with other distance measures, SSD shows superior discriminative ability for structural differences among different target categories.}}
        \label{fig:dist_comp}
    \end{figure}

    {The detailed calculation method for SSD is presented in Alg.~\ref{ag1}. Specifically, SSD is computed by solving for the minimum transportation cost based on EMD. In the SSD formulation, $f_{n,m}$ denotes the flow transported from the $n$-th point in $P_{i}$ to the $m$-th point in $Q_{j}$, while $w_n$ and $w_m$ represent the normalized weights (masses) assigned to the $n$-th source point and $m$-th target point, respectively. By optimizing the flows $f_{n,m}$ with respect to the sum constraints defined by $w_n$ and $w_m$, the SSD is computed as the resulting minimum transportation cost.
    To evaluate its effectiveness, Fig.~\ref{fig:dist_comp} compares the discriminability of different distance metrics across various target structures. Compared with traditional metrics such as L2 distance, cosine distance, and Hausdorff distance, the proposed SSD achieves a significantly lower distance distribution for intra-class target structure pairs (e.g., A220 vs. A220) and a higher separation from inter-class pairs (e.g., A220 vs. A330), resulting in less overlap between categories and greater inter-group discriminability. In addition, SSD yields a more compact and less dispersed intra-class distance distribution, indicating improved stability. These results highlight the effectiveness of SSD in handling discrete and uneven scattering points, and also demonstrate its clear advantages in distinguishing between different SAR target structures.}

    \begin{algorithm}
        \caption{Scattering Structure Distance.}
        \label{ag1}
        \begin{algorithmic}
            [1] \Statex \textbf{Input:} Region of interest $I_{0}$ and $I_{1}$. \Statex
            \textbf{Output:} Scattering structure distance $d_{ST}$. 
            \State {Apply Gaussian smoothing} to $I_{0}$ and $I_{1}$; 
            \State {Detect local maxima in $I_{0}$ and $I_{1}$ as scatter point sets:}
            $P = \{(x_{n}, y_{n}, a_{n})\}_{n=1}^{N}$, $Q = \{(x_{m}, y_{m}, a_{m}
            )\}_{m=1}^{M}$. Retain the top scatter points based on intensity $a$;
            \State {Compute the normalized weights for $P$ and $Q$:} $w_{n}= a_{n}/{\sum_{n=1}^{N}a_{n}}$,
            $w_{m}= a_{m}/{\sum_{m=1}^{M}a_{m}}$; 
            \State {Compute the scattering structure distance $d_{ST}$ between $P$ and $Q$ by EMD:}
            \[
                d_{ST}(P, Q) = \min \sum_{n=1}^{N}\sum_{m=1}^{M}f_{n,m}\cdot \sqrt{(x_{n}- x_{m})^{2}+ (y_{n}- y_{m})^{2}};
            \]
            \State {Constraints:}
            \[
                f_{n,m}\geq 0, \quad\sum_{n=1}^{N}\sum_{m=1}^{M}f_{n,m} = 1;
            \]
            \[
                {\sum_{m=1}^{M}f_{n,m}= w_{n}, \quad \forall n=1,\ldots,N;  }
            \]
            \[
                {\sum_{n=1}^{N}f_{n,m}= w_{m},\quad\forall m=1,\ldots,M;}
            \]
            \State {Solve the linear programming problem~\cite{sinkhorn} to compute $d_{ST}$:}
            \[
                d_{ST}= \frac{\sum_{n=1}^{N}\sum_{m=1}^{M}f_{n,m}\cdot \sqrt{(x_{n}- x_{m})^{2}+ (y_{n}- y_{m})^{2}}}{\sum_{n=1}^{N}\sum_{m=1}^{M}f_{n,m}}
                .
            \]
        \end{algorithmic}
    \end{algorithm}

    \begin{figure}[h]
        \centering
        \includegraphics[width=0.999\columnwidth]{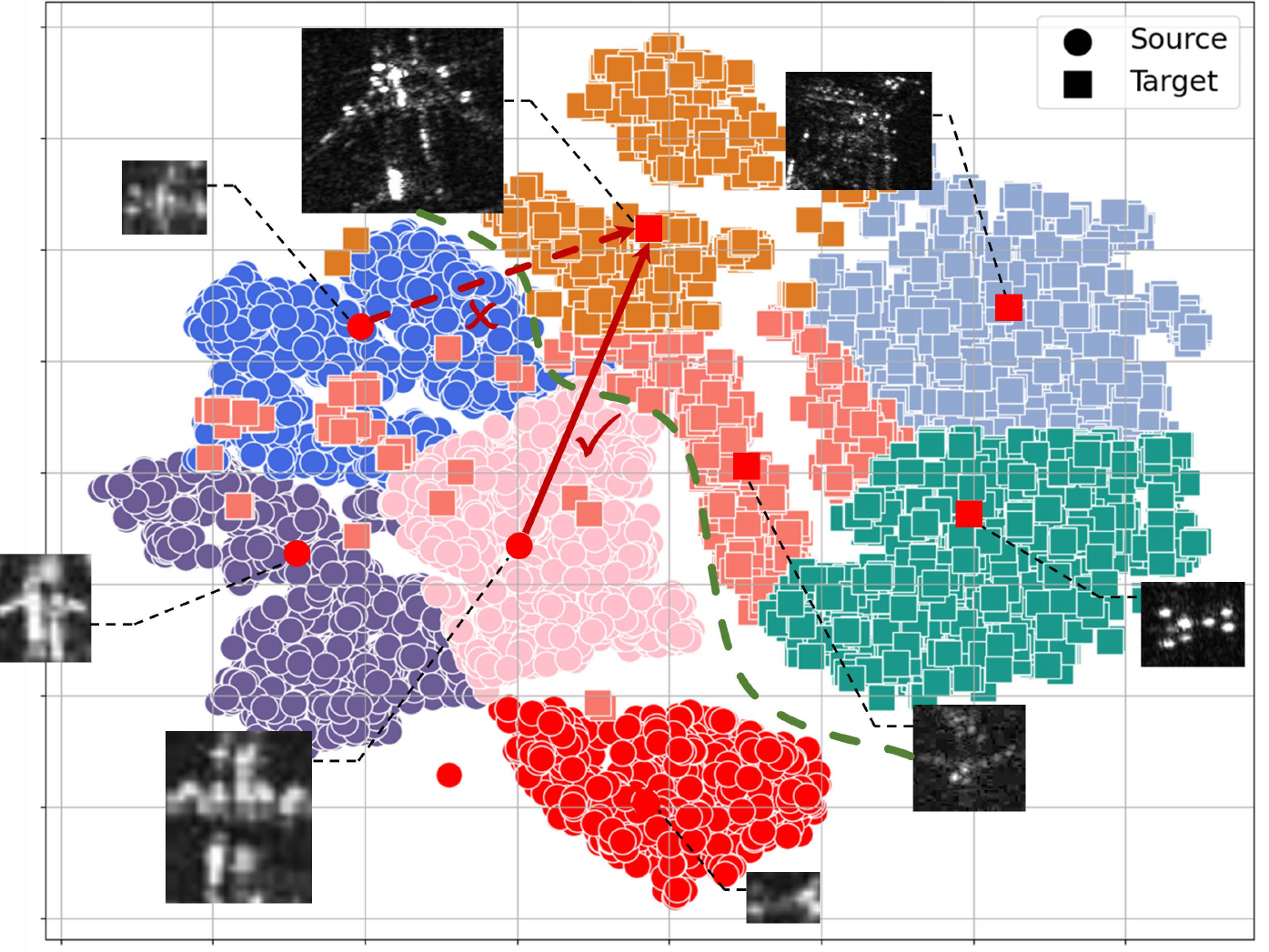}
        \caption{Feature distributions visualized by t-SNE~\cite{tsne} in the source and target domains. Clusters in different colors represent targets
        belonging to the same category but exhibiting different patterns due to structure variations. Directly aligning the overall feature distributions while disregarding intra-domain structural differences may result in misalignment between patterns with distinct structures, leading to negative transfer and hindering the achievement of optimal performance.}
        \label{fig:motivation}
        \vspace{-5mm}
    \end{figure}

    \subsection{Structure-induced Hierarchical Feature Adaptation}
    \label{SECSHFA}
    In the problem of cross-resolution domain adaptation, the data distributions between the low-resolution and high-resolution domains often exhibit multi-mode structures, as shown in the Fig~\ref{fig:motivation}. 
    Due to differences in imaging azimuth angles, target categories, and scales, SAR targets of the same category exhibit significant intra-class diversity. This phenomenon is further amplified across different resolutions, making it challenging to achieve satisfactory results using traditional domain adaptation methods that directly align overall distributions.
    If the intra-class diversity is ignored, errors may arise due to mismatched modes between different structures, leading to negative adaptation.

    To address this issue, \ourda is proposed to achieve structure-aware and fine-grained feature adaptation. In principle, the feature distributions of targets with similar structures should be more closely aligned. Thus, \ourda utilizes physical structure similarity to explicitly guide the feature adaptation process, ensuring that features with similar structural patterns are aligned in the feature space. 
    
\begin{figure}[!t]
        \centering
        \includegraphics[width=0.9999\columnwidth]{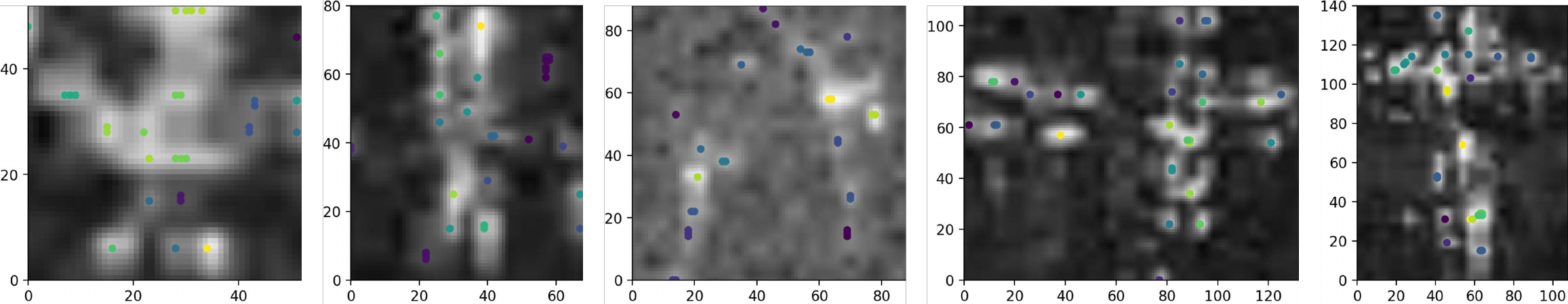}
        \caption{Examples of typical structures observed in the TerraSAR dataset ($n_{a}
        =5$). }
        \label{fig:typ_structure}
    \end{figure}

    Specifically, based on the scattering structure distance $d_{ST}$ introduced
    earlier, we use clustering methods to obtain $n_{a}$ structure anchors $\{A_{j}\}_{j=1}^{n_a}$, representing typical structures observed in the source domain data. 
    Examples of these typical structures are shown in the Fig.~\ref{fig:typ_structure}. 
    These typical structures reflect the main modes of the source domain feature distribution (visualized in Fig~\ref{fig:motivation}) and provide structural guidance for adapting domain features.
    Given the scattering point set $P_{i}$ of the $i\text{-th}$ instance, we can
    calculate the structural similarity $sim_{i,j}$ between this instance and
    each structure anchor $A_{j}$, which quantifies the relationship between the
    $i\text{-th}$ instance and the $j\text{-th}$ typical structures:
    \begin{equation}
        \label{eq10}
        {sim_{i,j}}= \max\left(\frac{{{e^{ - {d_{ST}}\left( {i,j} \right)}}}}{{\sum\nolimits_{j = 1}^{n_a} {{e^{ - {d_{ST}}\left( {i,j} \right)}}} }}, \delta\right).
    \end{equation}
    Here, $\delta=0.1$ is a margin parameter, used to keep structural similarity non-negative.
    $sim_{i,j}\in [\delta, 1]$. A higher $sim_{i,j}$ value indicates a greater structural similarity between instance $i$ and the structure anchor $A_{j}$. 
    Each structure anchor $A_{j}\in \{A_{j}\}_{j=1}^{n_a}$ is associated with a domain discriminator $\Disc{j}$, which is responsible for hierarchical feature adaptation corresponding to the structure $A_{j}$.
    By leveraging the structural similarity $sim_{i,j}$, we dynamically weight each instance feature to indicate how much the feature to be assigned to each discriminator, enabling structure-aware hierarchical feature adaptation:
    \begin{equation}
        \label{eq11}
        \begin{aligned}
            {\lossINS} & = - \frac{1}{n_a}\sum_{i \in \{n_a\}}\sum_{j \in \{|\src| \cup |\tgt|\}}\big[ y_{j}\log \Disc_{i}(sim_{i,j}\cdot f_{j}) \\
                       & + (1 - y_{j}) \log \big(1 - \Disc_{i}(sim_{i,j}\cdot f_{j})\big) \big],
        \end{aligned}
    \end{equation}
    where $y_{i}=1$ indicates the target domain and $y_{i}=0$ indicates the source domain. Combined with the traditional image-level adversarial loss ${\lossIMG}$ from~\cite{daf}, the feature adaptation loss is further expressed as:
    \begin{equation}
        \label{eq12}{\lossSHFA}={\lossIMG}+{\lossINS}.
    \end{equation}
    Here, the image-level feature adaptation loss ${\lossIMG}$ is defined as
    \begin{equation}
        \label{eq13}
        \begin{aligned}
            {\lossIMG}= -\sum_{j \in \{|\src| \cup |\tgt|\}} & \sum_{u, v}\big[ y_{j}\log \IMGDisc\big(f_{j}(u, v)\big)                  \\
            & + (1 - y_{j}\big) \log \big(1 - \IMGDisc\big(f_{j}(u, v)\big) \big) \big].
        \end{aligned}
    \end{equation}
    $\IMGDisc$ is the image-level domain discriminator. $f_{j}(u, v)$ is the
    image-level feature of the spatial location $(u,v)$.

    \vspace{-3mm}
    \subsection{Reliable Structural Adjacency Alignment}
\begin{figure}[!t]
        \centering
        \includegraphics[width=0.9999\columnwidth]{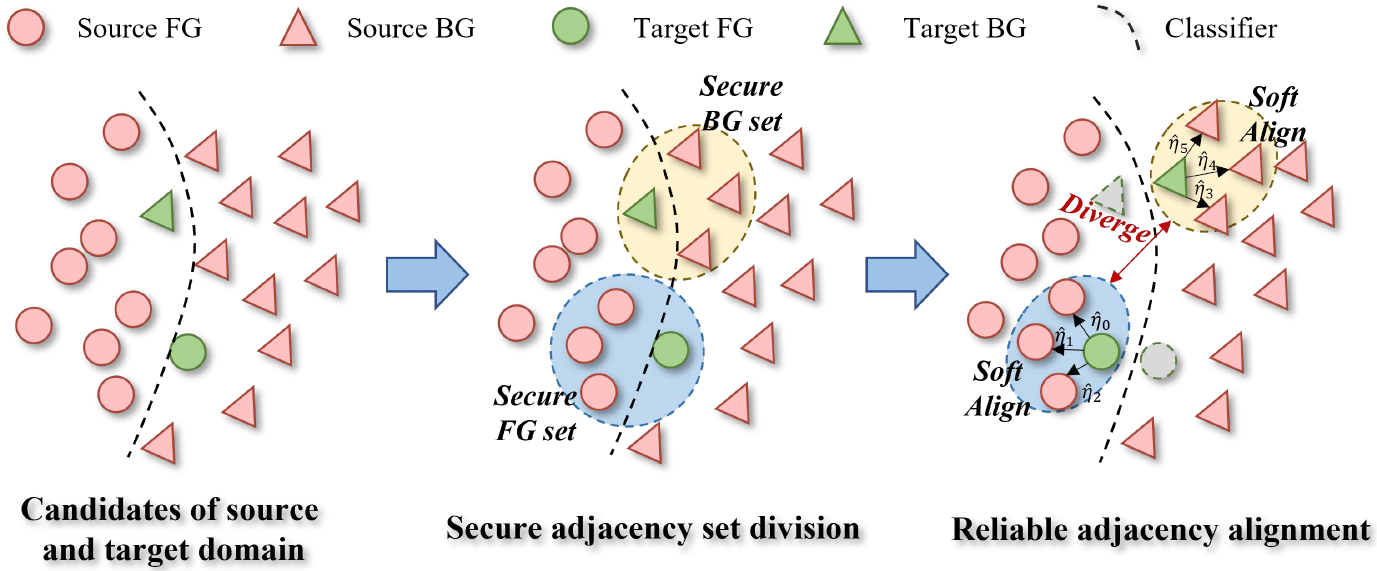}
        \caption{Illustration of \ouradja process. `FG' and `BG' denotes the abbreviation of foreground and background respectively. Aligning the target domain instance with its source domain neighbors enables the detector better discriminative capability in the target domain.}
        \label{fig:rsaa_process}
        \vspace{-5mm}
    \end{figure}

    The \ouradja~module is proposed to further enhance the discriminative capability of the detector in the target domain. The core idea of the \ouradja~module is that reliable neighbors in the feature space share the same semantics. Therefore, as shown in Fig.~\ref{fig:rsaa_process}, \ouradja~improves discriminability in the target domain by selecting the secure adjacency set from the source domain and aligning the semantics between target instances and the adjacency neighbors. Compared with traditional hard feature alignment methods, \ouradja~effectively alleviates the risky semantic propagation and error accumulation.
    
    Given a target domain instance $x^{T} \in \tgt$, we define the $r\text{-adjacency set}$ of $x^{T}$ as
    ${R_r}\left({{x^T}}\right) = \left\{{{x^T}}\right\} \cup \left\{{{N_r}\left( {{x^T}} \right)}
    \right\}$, where $N_{r}(x^{T})$ is the $r$ nearest source domain neighbors
    of the target domain instance $x^{T}$. Specifically:
    \begin{equation}
        \label{eq14}{N_r}\left({{x^T}}\right) = \left\{{x_j^S|j \in Top{\text{-}}r\left( {{\tgtfea} \otimes \{\srcfea_n\}_{n=1}^{n_f}} \right)}
        \right\},
    \end{equation}
    where $\otimes$ denotes cosine similarity. $\{\srcfea_n\}_{n=1}^{n_f}$ is the source feature bank $\{f_n^S,P_n^S,u_n^S, p_n^S\}_{n=1}^{n_f}$. $f_n^S$ and $P_n^S$ are the source instance feature and its scattering point set, respectively, while $u_n^S$ and $p_n^S$ are the predicted uncertainty and probability. 
    To measure the reliability of each instance $x^T$ and its adjacency set $R_{r}(x^{T})$, there are two factors defined: \ourinsfactor $\hat \eta$ and \ouradjafactor $\hat \gamma$.

    \subsubsection{\textit{Reliable Instance Factor} $\hat \eta$} 
     As introduced in Section~\ref{evidential modeling}, lower uncertainty $u$ indicates the prediction is  more reliable. Given the uncertainty $u(x_i^T)$ and its predicted probability $p_i^T$ of instance $x_i^T$, the \ourinsfactor $\hat \eta$ for a target domain instance is defined as:
    \begin{equation}
        \label{eq:k}
        \hat \eta \left( {x_i^T} \right) = p_i^T \cdot {e^{\left( { - u\left( {x_i^T} \right)/k} \right)}}.
    \end{equation}
    Here, $k$ is the scaling factor. When the uncertainty $u$ is lower, the \ourinsfactor $\hat{\eta}$ becomes higher, indicating that the prediction for the instance $x_i^T$ is more reliable. Such instance is more suitable to be assigned greater weights, while less reliable instances are penalized during semantic alignment. This helps to avoid error accumulation caused by unreliable predictions.
    
    \subsubsection{\textit{Secure Adjacency Factor} $\hat \gamma$} 
    It is used to select the secure adjacency set of each target domain instance $x^T$, making it suitable for adjacency semantic alignment.  
    We hypothesize that both structural and perceptual consistency are essential for a secure adjacency set. \textit{1) Structural consistency: instances within the secure adjacency set exhibit structural consistency. 2) Perceptual consistency: instances within the secure adjacency set have similar levels of reliability.}
    If both conditions are satisfied, we consider the adjacency set as a secure set, indicating the adjacency set has high-level structural and perceptual consistency. 
    Based on these two hypotheses, the secure adjacency factor $\hat \gamma \left( {x_i^T} \right)$ is defined from both structural and perceptual perspectives:
    \begin{equation}
        \begin{aligned}
            \hat \gamma \left( {x_i^T} \right) = \left\{ {{{\hat \gamma }_{ST}}\left( {x_i^T} \right),{{\hat \gamma }_u}\left( {x_i^T} \right)} \right\}.
        \end{aligned}
    \end{equation}
    Here,
    \begin{equation}
    \label{eq:securemaskmu}
        \begin{aligned}
        {{\hat \gamma }_{ST}}\left( {x_i^T} \right) &= \log \left( {1 + \frac{{\mathop {\max }\limits_{a \ne b,a,b \in \left| {{{\tilde D}_{ST}}} \right|} {{\left\| {{{\tilde D}_{ST}}\left( a \right) - {{\tilde D}_{ST}}\left( b \right)} \right\|}_1}}}{{\mathop {\min }\limits_{a \ne b} {{\left\| {{{\tilde D}_{ST}}\left( a \right) - {{\tilde D}_{ST}}\left( a \right)} \right\|}_1} + \epsilon }}} \right),\\
        {{\hat \gamma }_u}\left( {x_i^T} \right) &= \log \left( {1 + \frac{{\mathop {\max }\limits_{{x_i} \ne {x_j},{x_i},{x_j} \in {R_r}\left( {x_i^T} \right)} {{\left\| {u\left( {{x_i}} \right) - u\left( {{x_j}} \right)} \right\|}_1}}}{{\mathop {\min }\limits_{a \ne b} {{\left\| {u\left( {{x_i}} \right) - u\left( {{x_j}} \right)} \right\|}_1} + \epsilon }}} \right).
        \end{aligned}
    \end{equation}
    {Here, ${{\hat \gamma }_{ST}}\left( {x_i^T} \right)$ and ${{\hat \gamma }_{u}}\left( {x_i^T} \right)$ respectively characterize structural consistency and perceptual consistency for the $r\text{-adjacency set}$ of $x^{T}_i$.}
    ${\tilde D_{ST}} = \left\{ d_{ST}\left( i,j \right) \mid i,j \in |{R_r}\left( x_i^T \right)| \right\}$ denotes the set of scattering structure distances between all pairs of elements within the $r$-adjacency set ${R_r}\left( x_i^T \right)$. The uncertainty $u(x_i)$ quantifies the prediction reliability for the neighbors in the adjacency set.
    ${\left\|  \cdot  \right\|_1}$ is Manhattan distance. $\epsilon$ is a small constant to avoid division by zero.
    
    \subsubsection{Secure Adjacency Set Division}
    {Smaller values of ${{\hat \gamma }_{ST}}\left( x_i^T \right)$ and ${{\hat \gamma }_u}\left( x_i^T \right)$ indicate higher structural and perceptual consistency within the $r$-adjacency set $R_{r}(x_i^{T})$. Under such conditions, the adjacency set is categorized as a secure adjacency set $R_r^{se}$, which is used for reliable semantic alignment. In contrast, larger values of ${{\hat \gamma }_{ST}}\left( x_i^T \right)$ and ${{\hat \gamma }_u}\left( x_i^T \right)$ imply low consistency, and such risky adjacency sets are excluded from semantic alignment.}
  
    {Therefore, to ensure that the selected adjacency set satisfies both structural and perceptual consistency, we define secure adjacency set $R_r^{se}$ as the intersection of the structurally consistent set $R_r^{ST}$ and the perceptually consistent set $R_r^u$:}
    \begin{equation}
        \label{eq17}
        R_r^{se} = R_r^{ST} \cap R_r^u.
    \end{equation}
    {Here, $R_r^{ST}$ and $R_r^u$ are selected from $R_r$ by applying the structural consistency constraint ${{\hat \gamma }_{ST}}\left( x_i^T \right)$ and the perceptual consistency constraint ${{\hat \gamma }_u}\left( x_i^T \right)$, respectively. To enhance the flexibility of the selection process, we adopt an adaptive selection strategy:}
    \begin{equation}
        \label{eq18}
        \begin{aligned}
        R_r^{ST} &= \left\{ {{R_r}\left( {x_i^T} \right)|{{\hat \gamma }_{ST}}\left( {x_i^T} \right) \le {\mu _{ST}} + {\lambda _{se}} \cdot {\sigma _{ST}}} \right\},\\
        R_r^u &= \left\{ {{R_r}\left( {x_i^T} \right)|{{\hat \gamma }_u}\left( {x_i^T} \right) \le {\mu _u} + {\lambda _{se}} \cdot {\sigma _u}} \right\}.
        \end{aligned}
    \end{equation}
    {$\mu_{ST}$ and $\sigma_{ST}$ are the mean and standard deviation of $\hat{\gamma}_{ST}$ over the training batch. Similarly, $\mu_u$ and $\sigma_u$ are defined for $\hat{\gamma}_u$.  $\lambdaSecmask$ is a scaling hyperparameter. 
    Following this strategy, all $r$-adjacency set $R_r(x_i^T)$ of $x_i^T \in \tgt$ that meet the meet the constraints in Eq.\ref{eq18} are included in the secure adjacency set $R_r^{se}$.}
    
    \subsubsection{Soft Semantic Alignment}
    {To enhance the detector’s discriminability in the target domain, we use the \ourinsfactor $\hat \eta$ and the secure adjacency set $R_r^{se}$ to guide soft semantic alignment, thereby transferring semantic information from the source domain to the target domain. Furthermore, we encourage the separation of feature distributions between foreground and background instances to further improve the model's discriminability. Specifically, the secure adjacency set $R_r^{se}$ is partitioned as secure foreground adjacency set $R_r^{fg}$ and secure background adjacency set $R_r^{bg}$ according to the probabilities $p_j$ of neighbor instances $x_j \in N_{r}(x_i^{T})$:}
    \begin{equation}
        \label{eqbgfg}
        \begin{aligned}
        R_r^{fg} &= \left\{ {R_r^{se}\left( {x_i^T} \right)|\mathop {\min }\limits_{x_j \in {N_r}\left( {x_i^T} \right)} p_j \ge 0.5} \right\},       \\
        R_r^{bg} &= \left\{ {R_r^{se}\left( {x_i^T} \right)|\mathop {\min }\limits_{x_j \in {N_r}\left( {x_i^T} \right)} p_j < 0.5} \right\}.
        \end{aligned}
    \end{equation}
    Afterwards, \ourinsfactor together with the secure adjacency sets $R_r^{fg}$ and $R_r^{bg}$ are employed for soft semantic alignment, which encourages the same semantic features to be closer and those of different semantics to be separated:
    \begin{equation}
        \label{eqrsaa}
        \begin{aligned}
        {\lossRSAA} &= \frac{1}{{\left| {R_r^{se}} \right|}}\sum\limits_{x_i^T \in R_r^{se}} {\sum\limits_{x_j^S \in {N_r}\left( {x_i^T} \right)} {\hat \eta \left( {x_i^T} \right){{\left\| {f_i^T - f_j^S} \right\|}_1}} } \\
         &+ \frac{1}{{\left| {R_r^{fg}} \right|\left| {R_r^{bg}} \right|}}\sum\limits_{x_i^T \in R_r^{fg}} {\sum\limits_{x_j^T \in R_r^{bg}} {\max \left( {\varsigma  - {{\left\| {\bar f_i^{fg} - \bar f_j^{bg}} \right\|}_1},0} \right)}}.
        \end{aligned}
    \end{equation}
    $\varsigma=0.2$ is the margin parameter.  $\bar f_i^{fg}$ and $\bar f_j^{bg}$ represent the average foreground feature  and the average background feature respectively within the secure set. By minimizing ${\lossRSAA}$, the target domain instance $x^{T}$ gains discriminative knowledge from source reliable neighbors in the secure set $R_r^{se}$, thereby  enhancing the performance of the detector in the target domain.
    
    \subsection{Overall Training Objective}
    In summary, to achieve cross-resolution SAR target detection, the overall optimization objective of \ourmethod is expressed as:
    \begin{equation}
        \label{eq:loss}
        \begin{aligned}
            {\loss}_{\text{overall}}= \lossSRC+ {\lossEVI}+{\lambdaSHFA}{\lossSHFA}+{\lambdaRSAA}{\lossRSAA},
        \end{aligned}
    \end{equation}
    where $\lambdaSHFA$ and $\lambdaRSAA$ are hyper-parameters that  balance the contributions of each component. The training process is detailed in Alg.~\ref{alg:2}.

    \begin{algorithm}
        \caption{Training process of \ourmethod.}
        \label{alg:2}
        \begin{algorithmic}
            [1] \Statex \textbf{Input:} Labeled source domain data
            $\{x^{S}, y^{S}\}$ and unlabeled target domain data $x^{T}$, feature extractor $\backbone$,
            detection branch $\dethead$, evidential branch $\evihead$, domain
            discriminator $\Disc$, training epochs $N_{\mathrm{epoch}}$.

            \Statex \textbf{Output:} Well-trained $\backbone$, $\dethead$.

            \State Initialize parameters of $\backbone$, $\dethead$, $\evihead$, and $\Disc$. 
            \State Construct scattering structure anchors $\{A_{j}
            ^{n}\}_{j=1}^{a}$ from $\{x^{S}, y^{S}\}$. \For{$n = 1$ to $N_{\mathrm{epoch}}$}

            \State Calculate ${\lossEVI}$ and $\lossSRC$ by Eq.~\ref{eqsrc} using $x^{S}$
            and $y^{S}$; 
            \State
            {Take gradient descent $\nabla_{\{\backbone,\evihead\}}{\lossEVI}+ \nabla_{\{\backbone,\dethead\}}{\loss}_{S}$};
            \State Update $\backbone$, $\evihead$ and $\dethead$; \State Calculate
            structure similarity by Alg.~\ref{ag1} and Eq.~\ref{eq10}; \State
            Calculate ${\lossSHFA}$ by Eq.~\ref{eq12}; \State
            {Take gradient descent $\nabla_{\{\Disc\}}{\lossSHFA}- \nabla_{\{\backbone\}}{\lossSHFA}$};
            \State Update $\backbone$ and $\Disc$; \If{{$n > N_{\mathrm{epoch}}/2$}}
            \State Generate $r$-adjacency of $x^{T}$; \State Compute
            ${\lossRSAA}$ using Eq.~\ref{eqrsaa}; \State
            {Take gradient descent $\nabla_{\{\backbone, \dethead\}}{\lossRSAA}$};
            \State Update $\backbone$ and $\dethead$;

            \EndIf \EndFor \State \textbf{return} {$\backbone$ and $\dethead$.}
        \end{algorithmic}
    \end{algorithm}

    \section{Experiments and Analysis}
    \label{Experiments}
    \subsection{Dateset Descriptions}
    In this experiment, SAR images with different resolutions are used to construct cross-resolution SAR target detection tasks. 
    The specific data configurations are shown in Table~\ref{tab:data}. 
    The experimental dataset includes Gaofen-3 aircraft data (1 m resolution) from~\cite{AIRcraft-1.0}, TerraSAR aircraft data (3 m resolution) from~\cite{zhang2022sefepnet}, and FARAD vehicle data (0.5 m and 0.1 m resolutions) from~\cite{FARAD,zou2023cross}.
    During the data preprocessing stage, to establish a unified cross-resolution target detection benchmark, we first used a bilinear interpolation algorithm to up-sample the low-resolution images to match the sizes of the corresponding high-resolution images. Subsequently, the images were cropped into $256\times256$ image patches. 
    Based on the various resolution SAR data shown in Table~\ref{tab:data} and adhering to the principle of consistency in target categories between the source and target domains, we constructed four cross-resolution target detection  tasks to verify the effectiveness of the proposed method. 
    The cross-resolution detection task configurations are detailed in Table~\ref{tab:task}.

    \begin{table}[t]
        \caption{Technical parameters of different resolution SAR sensors.}
        \label{tab:data}
        \centering
        \setlength{\tabcolsep}{10pt}
        { \resizebox{1\columnwidth}{!}{ \renewcommand{\arraystretch}{1.} \begin{tabular}{@{}ccccc@{}}\toprule Sensor & {Gaofen-3} & {TerraSAR} & {FARAD (HR)} & {FARAD (LR)}\\ \midrule Imaging platform & {Spaceborne} & {Spaceborne} & {Airborne} & {Airborne} \\ Band & {C} & {X} & {Ka} & {Ka}\\ Resolution & {$1\text{m}\times1$m} & {$3\text{m}\times3$m} & {$0.1\text{m}\times0.1$m} & {$0.5\text{m}\times0.5$m}\\ Target type & {Aircraft} & {Aircraft} & {Vehicle} &{Vehicle} \\ Slice number & {3500} & {2674} & {1265} &{1129} \\ \bottomrule\end{tabular}%
        }}
    \end{table}

    \begin{table}[t]
        \caption{Cross-resolution detection tasks in the experiments. `LR' denotes low resolution. `HR' denotes high resolution.}
        \label{tab:task}
        \centering
        \setlength{\tabcolsep}{10pt}
        { \resizebox{1\columnwidth}{!}{ \renewcommand{\arraystretch}{1.} \begin{tabular}{@{}ccc@{}}\toprule Task & {Source Domain Data} & {Target Domain Data} \\ \midrule Aircraft LR$\rightarrow$HR & {TerraSAR ($3\text{m}\times3$m)} & {Gaofen-3 ($1\text{m}\times1$m)} \\ 
        Aircraft HR$\rightarrow$LR & {Gaofen-3 ($1\text{m}\times1$m)} & {TerraSAR ($3\text{m}\times3$m)} \\ 
        Vehicle LR$\rightarrow$HR & {FARAD ($0.5\text{m}\times0.5$m)} & {FARAD ($0.1\text{m}\times0.1$m)} \\ 
        Vehicle HR$\rightarrow$LR & {FARAD ($0.1\text{m}\times0.1$m)} & {FARAD ($0.5\text{m}\times0.5$m)} \\ \bottomrule\end{tabular}
        }}
    \end{table}

    \subsection{Implementation Details}
    All experiments were conducted on a server equipped with an NVIDIA A6000 GPU with 48GB of memory. During the training phase, labeled source domain data and unlabeled target domain data were jointly input into the target detection model for training 10 epochs. The initial learning rate was set as $0.002$, and decreased to $0.001$ after 5 epochs. In the testing phase, the target domain data were used to evaluate the detection performance. The evaluation metrics we adopted include VOC mAP, F1 score, Precision, and Recall. In all experiments, the IoU threshold for calculating mAP was set to 0.5.
    Hyper-parameter $\lambdaSHFA=0.5$,  $\lambdaRSAA=0.5$, $\lambdaSecmask=-1$, $k=30$ $n_a=5$, $r=5$ for Aircraft tasks, and $r=3$ for Vehicle tasks. Details are discussed in Section~\ref{discuss}.

    \subsection{Comparisons and Analysis}

    {In the comparative experiments, we compared seven SOTA domain adaptation methods, including Domain Adaptive Faster RCNN~\cite{daf} (DAF), Graph-induced Prototype Alignment Network~\cite{gpa} (GPANet), Imbalanced Discriminant Alignment~\cite{ida} (IDA), Hierarchical Similarity Alignment Neural Network~\cite{HSA} (HSANet), Foreground Instance Enhancement Network~\cite{fie} (FIENet), Class-Aware Teacher~\cite{cat} (CAT), and Masked Image Consistency~\cite{mic} (MIC). Among these methods, only FIENet is based on the FCOS~\cite{fcos} detection model, while all other methods use the Faster RCNN detector with ResNet-101. To ensure fair comparison, all experiments were conducted with the same input image size and identical data augmentation strategies. Comparative experiments were carried out on the four cross-resolution target detection tasks presented in Table~\ref{tab:task}}

    \subsubsection{Quantitative comparisons}
    From the quantitative results in Table~\ref{tab:aircraft} and Table~\ref{tab:vehicle}, it is clear that the performance of Faster RCNN without domain adaptation is significantly lower than that of other methods. 
    This demonstrates that discrepancies across resolutions can lead to a substantial decline in SAR target detection performance. 
    Moreover, domain adaptive detection methods consistently outperform the baseline Faster RCNN, thus validating the feasibility and effectiveness of domain adaptation-based approaches for cross-domain target detection tasks.

    {According to the meter-level cross-resolution results in Table~\ref{tab:aircraft}, HSANet and DAF perform poorly in the Aircraft LR→HR and Aircraft HR→LR tasks. 
    This is because these cross-resolution detection tasks involve a greater diversity in target scales, which increases the challenge of cross-resolution target detection. 
    In contrast, the proposed \ourmethod outperforms the other detection methods across all evaluation metrics. This not only highlights the effectiveness of \ourmethod in cross-resolution detection, but also further demonstrates its robustness and performance advantages in addressing resolution distribution discrepancies.
    For the meter-level cross-resolution Aircraft LR→HR and HR→LR tasks, \ourmethod achieves improvements of more than 14~\% in Recall, mAP, and F1 score, and approximately 4~\% in Precision, compared with the second-best method.
    These results indicate that our method significantly enhances the detector’s transferability and discriminability for the cross-resolution detection task. This is achieved by aligning feature distributions effectively by \ourda and transferring high-quality, discriminative semantics via \ouradja.}

    {As for the sub-meter level cross-resolution results (Vehicle LR→HR and Vehicle HR→LR) in Table~\ref{tab:vehicle}, the mAP scores are closer to the F1 scores. This can be attributed to the relatively uniform scale distribution of vehicles in the dataset, which reduces the challenge of bounding box regression during the detection process. Nevertheless, our method still achieves the best performance in Recall, mAP, and F1 in both tasks.
    Specifically, in the Vehicle HR→LR task, our method achieves 0.652 mAP and 0.701 F1 score, outperforming the second-best method, MIC, by approximately 16~\%.
    Although, in the Vehicle LR→HR task, our method's Precision is slightly lower than IDA by 1.2~\%, it surpasses IDA by at least 10~\% in Recall, F1, and mAP metrics, indicating that our method achieves a better balance between recall and precision.}
    
    {Based on the above results and analysis, our method is not only suitable for addressing resolution discrepancies in meter-level datasets, but is also effective for cross-resolution target detection in sub-meter, very high-resolution datasets.}

\begin{table}[t]
    \centering
    \caption{{Statistic comparisons with SOTA UDA methods on
    aircraft Tasks. The bold figure denotes the best score. The underline indicates the second-best performance. The symbol `*' denotes the detector based on the FCOS model.}}
    \label{tab:aircraft}
    \resizebox{\columnwidth}{!}{%
    \begin{tabular}{@{}c|c|cccc@{}}
    \toprule
    Task                              & Method      & Recall          & Precision       & mAP              & F1              \\ \midrule
    \multirow{9}{*}{Aircraft   LR→HR} 
    & Faster RCNN~\cite{fasterrcnn}  & 0.171            & 0.280            & 0.145             & 0.211            \\
         & DAF~\cite{daf}         & 0.284          & 0.461          & 0.239          & 0.351          \\
         & GPANet~\cite{gpa}      & \underline{0.494}          & {0.563}          & \underline{0.490}           & \underline{0.526}          \\
         & IDA~\cite{ida}         & 0.257          & \underline{0.688}          & 0.201            & 0.375          \\
         & HSANet~\cite{HSA}      & 0.282          & 0.438          & 0.193           & 0.343          \\
         & {FIENet*~\cite{fie}}   & {0.482}       & {0.378}      & {0.355}                      & {0.423}          \\
         & {CAT~\cite{cat}}       & {0.364}       & {0.437}      & {0.301}                      & {0.397}          \\
         & {MIC~\cite{mic}}       & {0.485}       & {0.554}    & {0.386}                     & {{0.517}}     \\
              
          & Ours method       & \textbf{0.658}   & \textbf{0.722}   & \textbf{0.563}    & \textbf{0.688}   \\ \midrule
          
    \multirow{9}{*}{Aircraft   HR→LR} & Faster RCNN~\cite{fasterrcnn} & 0.326           & 0.437           & 0.240            & 0.373           \\
          & DAF~\cite{daf}         & 0.436          & 0.428      & 0.308           & 0.432          \\
          & GPANet~\cite{gpa}      & \underline{0.495}           & {0.601}         & \underline{0.448}     & {0.543}          \\
          & IDA~\cite{ida}         & 0.392          & 0.598          & 0.339           & 0.473          \\
          & HSANet~\cite{HSA}      & 0.362          & 0.578          & 0.302           & 0.445          \\

          & {FIENet*~\cite{fie}}       & {0.461}          & {0.354}        & {0.328}          & {0.401}          \\
          & {CAT~\cite{cat}}       & {0.359}          & {0.474}           & {0.331}          & {0.409}          \\
          & {MIC~\cite{mic}}       & {0.490}   & {\underline{0.657}}    & {0.402}    & {\underline{0.561}}     \\
          & Ours method       & \textbf{0.657} & \textbf{0.685} & \textbf{0.524} & \textbf{0.671} \\ \bottomrule
    \end{tabular} %
    }
    \end{table}

\begin{table}[t]
    \centering
    \caption{{Statistic comparisons with SOTA UDA methods on
    vehicle Tasks. The bold figure denotes the best score. The underline indicates the second-best performance. The symbol `*' denotes the detector based on the FCOS model.}}
    \label{tab:vehicle}
    \resizebox{\columnwidth}{!}{%
    \begin{tabular}{@{}c|c|cccc@{}}
    \toprule
    Task                           & Method      & Recall          & Precision       & mAP             & F1              \\ \midrule
    \multirow{9}{*}{Vehicle LR→HR} 
        & Faster RCNN~\cite{fasterrcnn} & 0.583      & 0.449           & 0.578           & 0.574           \\
               & DAF~\cite{daf}         & 0.584          & 0.640         & 0.597          & 0.611          \\
               & GPANet~\cite{gpa}       & 0.604          & 0.617          & 0.605          & 0.611          \\
               & IDA~\cite{ida}         & {0.607}         & \textbf{0.684}          & \underline{0.629}          & {0.643}          \\
               & HSANet~\cite{HSA}       & 0.669          & 0.623         & 0.597          & 0.645          \\
               & {FIENet*~\cite{fie}}       & {0.610}          & {0.600}        & {0.606}          & {0.605}          \\
               & {CAT~\cite{cat}}       & {{0.638}}          & {0.628}           & {0.607}          & {0.633}          \\
              & {MIC~\cite{mic}}       & {\underline{0.670}}   & {{0.629}}    & {0.603}    & {\underline{0.649}}     \\
               & Ours method       & \textbf{0.723} & \underline{0.676} & \textbf{0.718} & \textbf{0.713} \\ \midrule
               
    \multirow{9}{*}{Vehicle HR→LR} 
    & Faster RCNN~\cite{fasterrcnn} & 0.337           & 0.545           & 0.271           & 0.416           \\
           & DAF~\cite{daf}        & 0.479          & 0.483          & 0.420          & 0.481          \\
           & GPANet~\cite{gpa}      & 0.525          & 0.513          & 0.430         & 0.519          \\
           & IDA~\cite{ida}  & \underline{0.559}    & 0.555         & {0.503}        & {0.557}          \\
           & HSANet~\cite{HSA}       & 0.472          & {0.606}      & 0.405          & 0.530          \\
          & {FIENet*~\cite{fie}}       & {0.510}          & {0.479}       & {0.430}          & {0.494}          \\
          & {CAT~\cite{cat}}       & {0.541}         & {0.605}           & {0.542}          & {0.571}          \\
          & {MIC~\cite{mic}}       & {0.550}          & {\underline{0.660}}           & {\underline{0.553}}          & {\underline{0.601}}          \\
           & Ours method       & \textbf{0.706} & \textbf{0.695} & \textbf{0.652} & \textbf{0.701} \\ \bottomrule
    \end{tabular}%
    }
    \end{table}

    \subsubsection{Qualitative comparisons}
    \begin{figure}[h]
        \centering
        \includegraphics[width=0.9999\columnwidth]{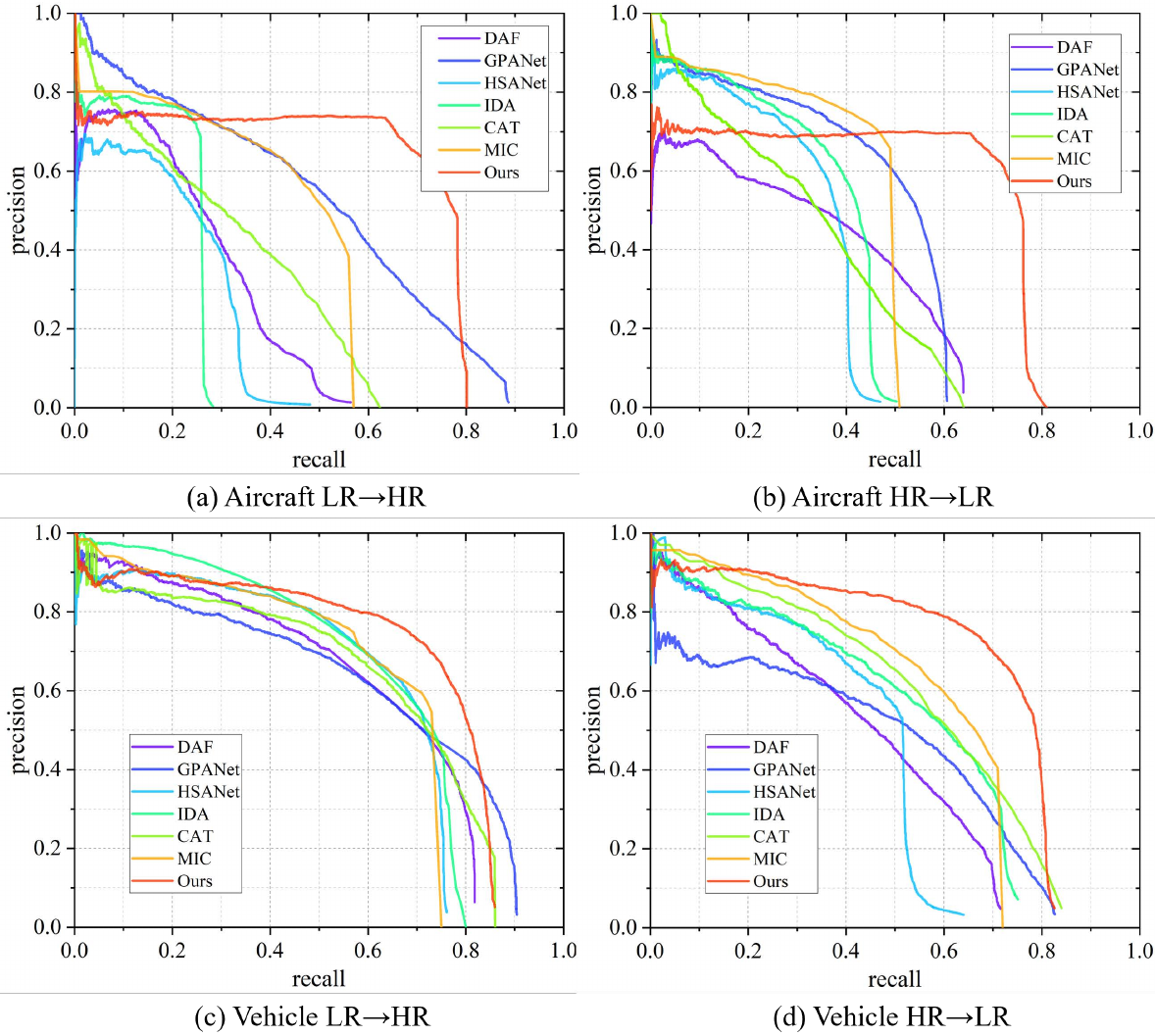}
        \caption{{Precision-Recall (PR) curves of comparison methods.}}
        \label{fig:pr}
    \end{figure}
    The Precision-Recall (PR) curves of comparison methods across different cross-domain detection tasks are shown in Fig.~\ref{fig:pr}. The PR curves show that the area under the curve (AUC) of our method is significantly larger across all tasks, demonstrating its superiority. 
    Additionally, the PR curve of the proposed method is more stable compared to other methods, indicating that the detection precision remains consistent under varying recall rates, highlighting stronger generalization capabilities. 
    
\begin{figure*}[h]
        \centering
        \includegraphics[width=0.9999\textwidth]{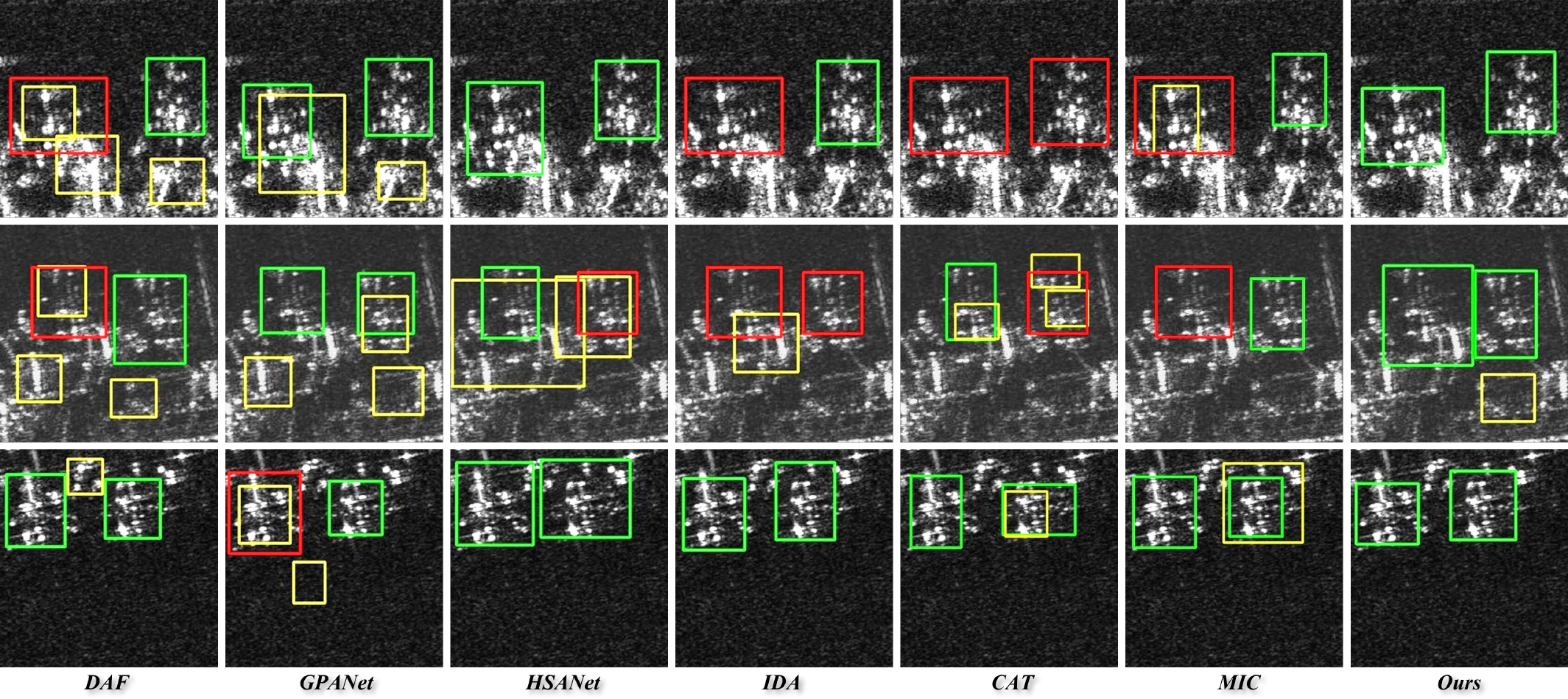}
        \caption{{Detection visualizations of different methods on the Aircraft LR$\rightarrow$HR task. Here, in all subfigures, the green, yellow and red rectangular boxes denote correct detections, false alarms and miss detections respectively.}}
        \label{fig:aircraft_lr2hr}
    \end{figure*}

\begin{figure*}[!t]
        \centering
        \includegraphics[width=0.9999\textwidth]{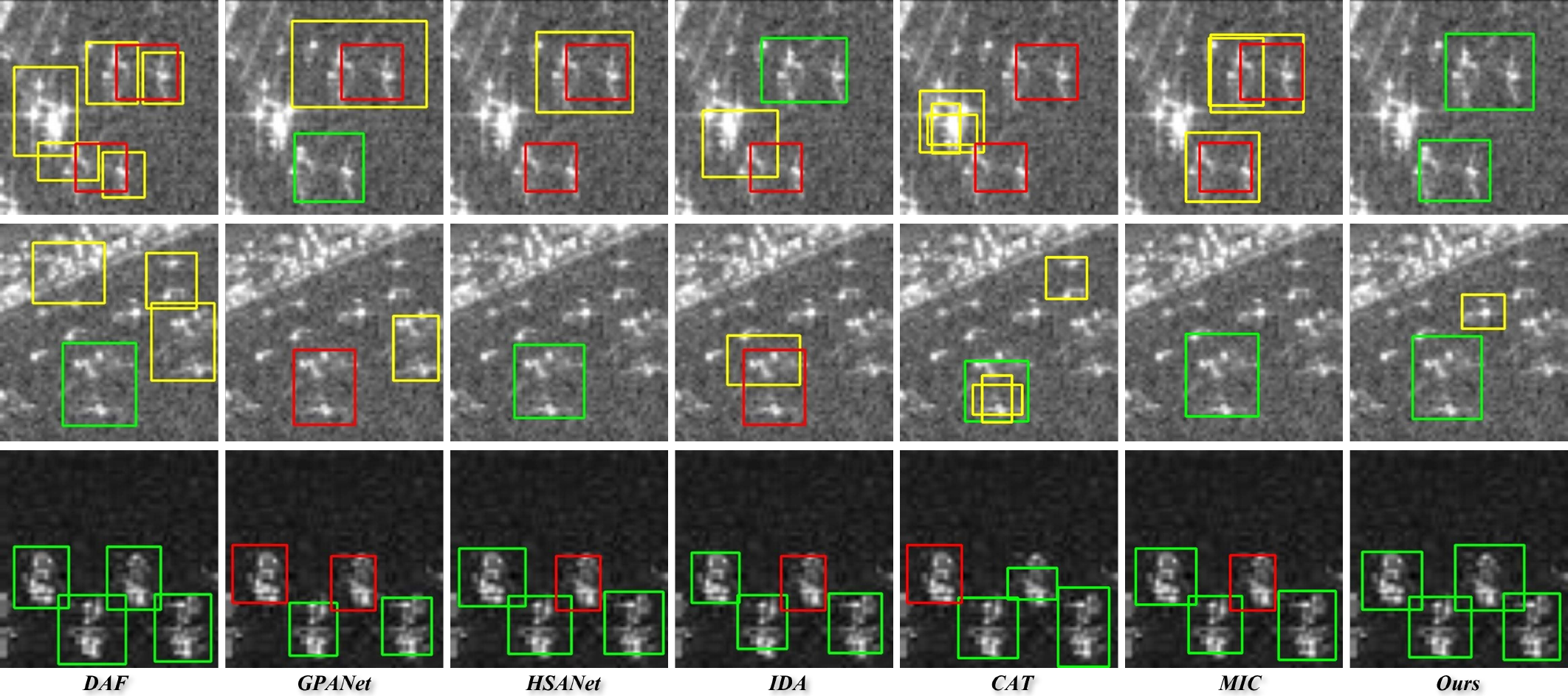}
        \caption{{Detection visualizations of different methods on the Aircraft HR$\rightarrow$LR task. Here, in all subfigures, the green, yellow and red rectangular boxes denote correct detections, false alarms and miss detections respectively.}}
        \label{fig:aircraft_hr2lr}
    \end{figure*}

\begin{figure*}[!t]
        \centering
        \includegraphics[width=0.9999\textwidth]{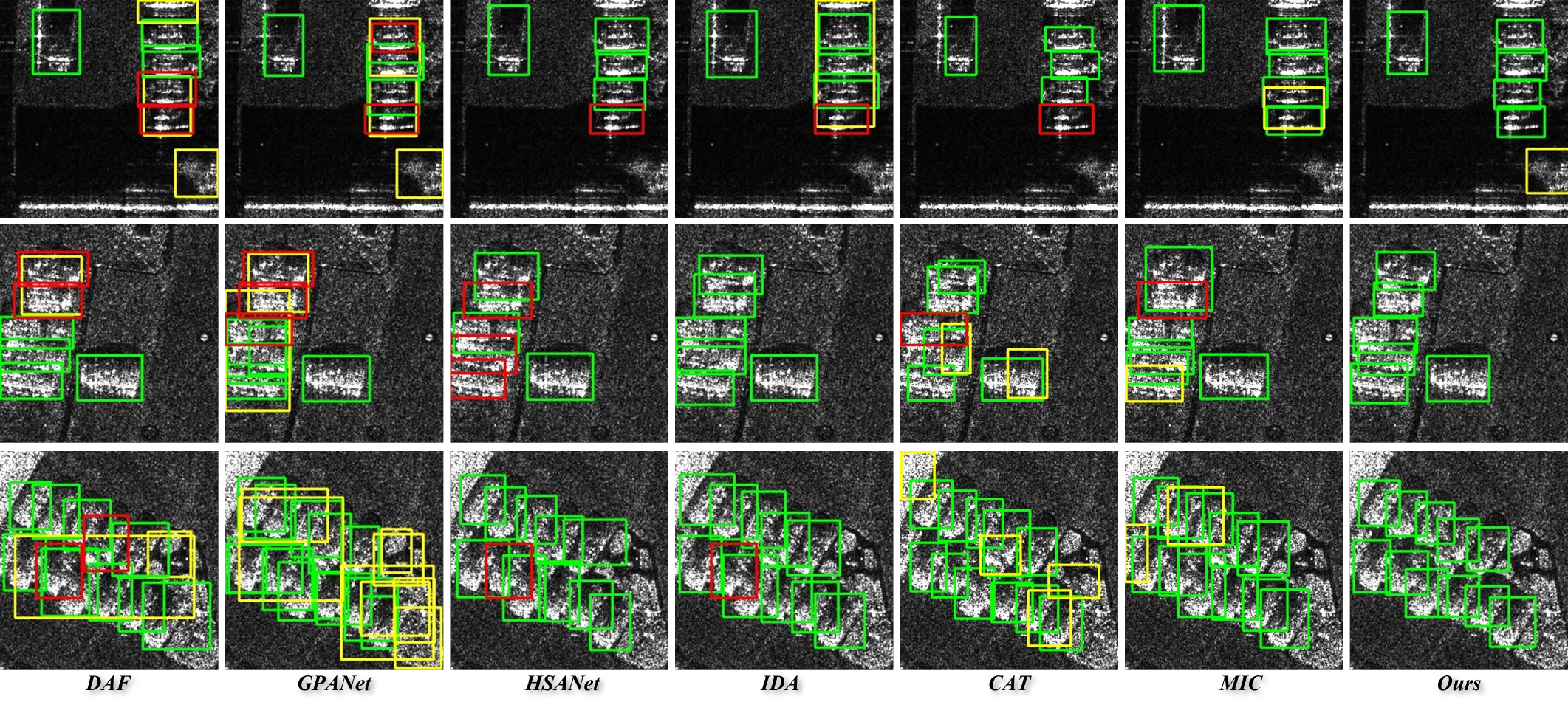}
        \caption{{Detection visualizations of different methods on the Vehicle LR$\rightarrow$HR task. Here, in all subfigures, the green, yellow and red rectangular boxes denote correct detections, false alarms and miss detections respectively.}}
        \label{fig:farad_lr2lhr}
    \end{figure*}

\begin{figure*}[t]
        \centering
        \includegraphics[width=0.9999\textwidth]{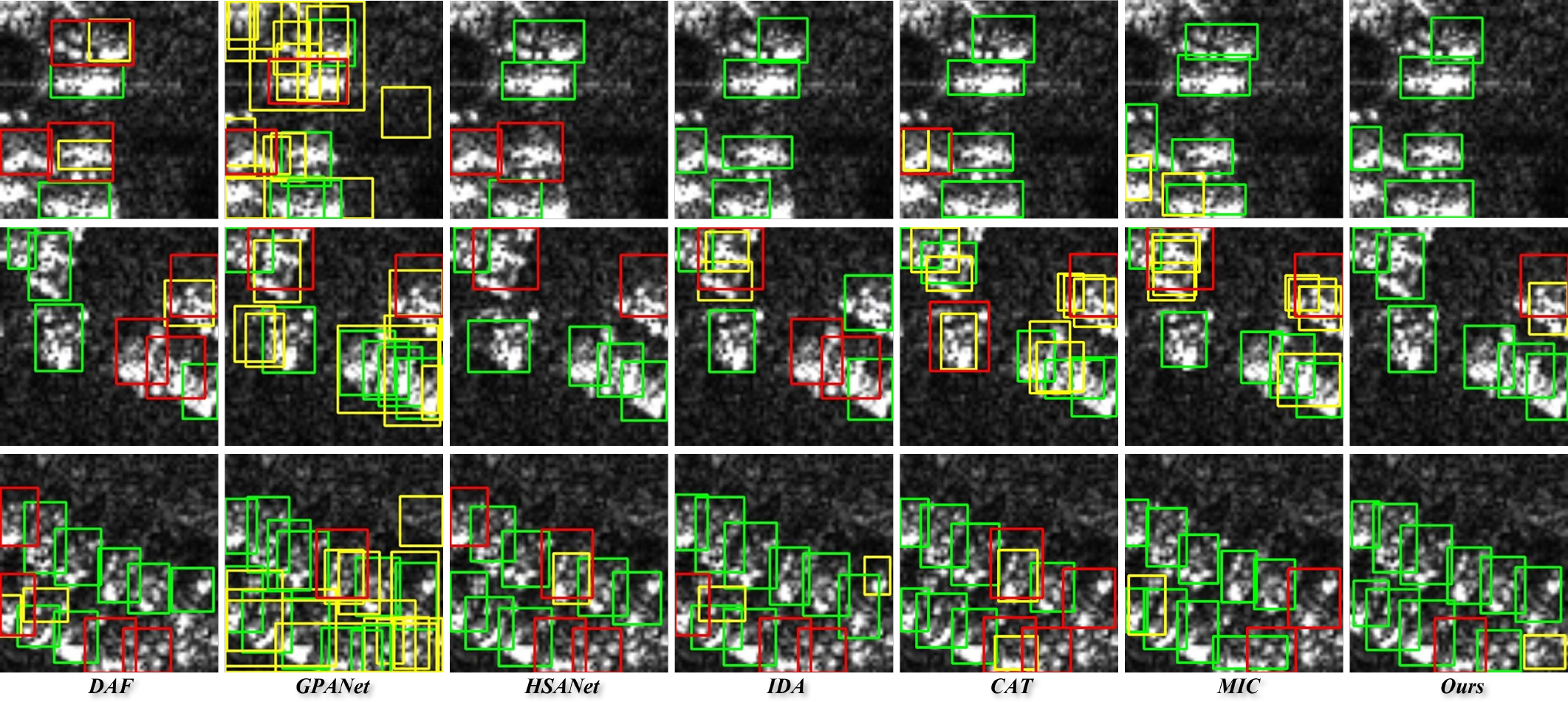}
        \caption{{Detection visualizations of different methods on the Vehicle HR$\rightarrow$LR task. Here, in all subfigures, the green, yellow and red rectangular boxes denote correct detections, false alarms and miss detections respectively.}}
        \label{fig:farad_hr2lr}
    \end{figure*}

    Moreover, to provide an intuitive comparison, the visualization results of detection are shown in Fig~\ref{fig:aircraft_lr2hr}, Fig~\ref{fig:aircraft_hr2lr}, Fig~\ref{fig:farad_lr2lhr}, and Fig~\ref{fig:farad_hr2lr}. 
    From these visualizations, although the other comparison methods can detect most targets, they also exhibit a large number of false positives and missed detections. 
    In contrast, the proposed method significantly reduces false positives and missed detections. 
    This improvement is particularly evident in complex scenarios, where the proposed method demonstrates stronger robustness and adaptability.
    From the aircraft detection results shown in Fig.~\ref{fig:aircraft_lr2hr}, our method demonstrates remarkable discriminative performance even in complex contextual backgrounds, such as near jet bridges. {Compared with other methods that rely solely on adversarial feature alignment, our method enhances the discriminability by leveraging reliable source discrimination knowledge transfer through \ouradja.} This improvement enables our method to achieve superior performance in complex background scenarios.
    
    \begin{figure}[!h]
        \centering
        \includegraphics[width=0.9999\columnwidth]{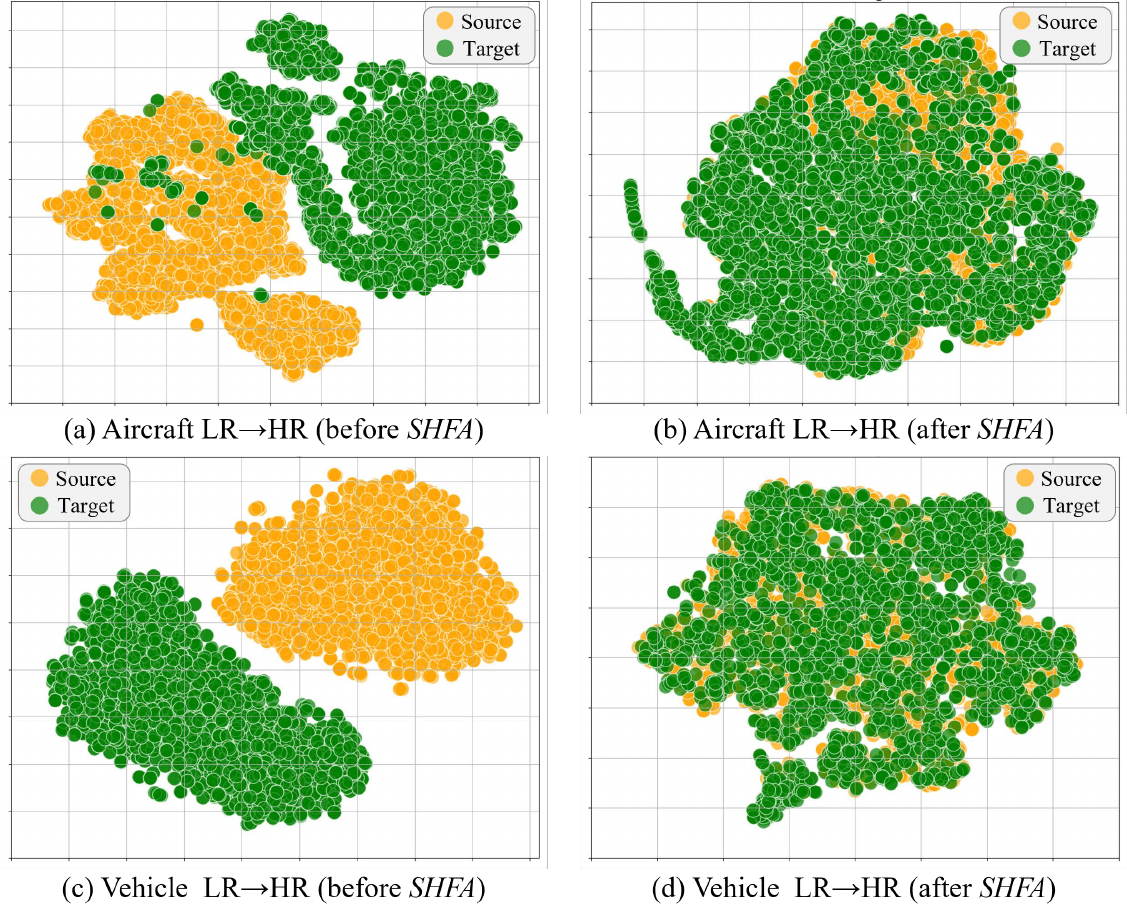}
        \caption{Feature visualization by t-SNE~\cite{tsne} of instance features of source and target domain.}
        \label{fig:tsne}
    \end{figure}

    \begin{figure}[h]
        \centering
        \includegraphics[width=0.999\columnwidth]{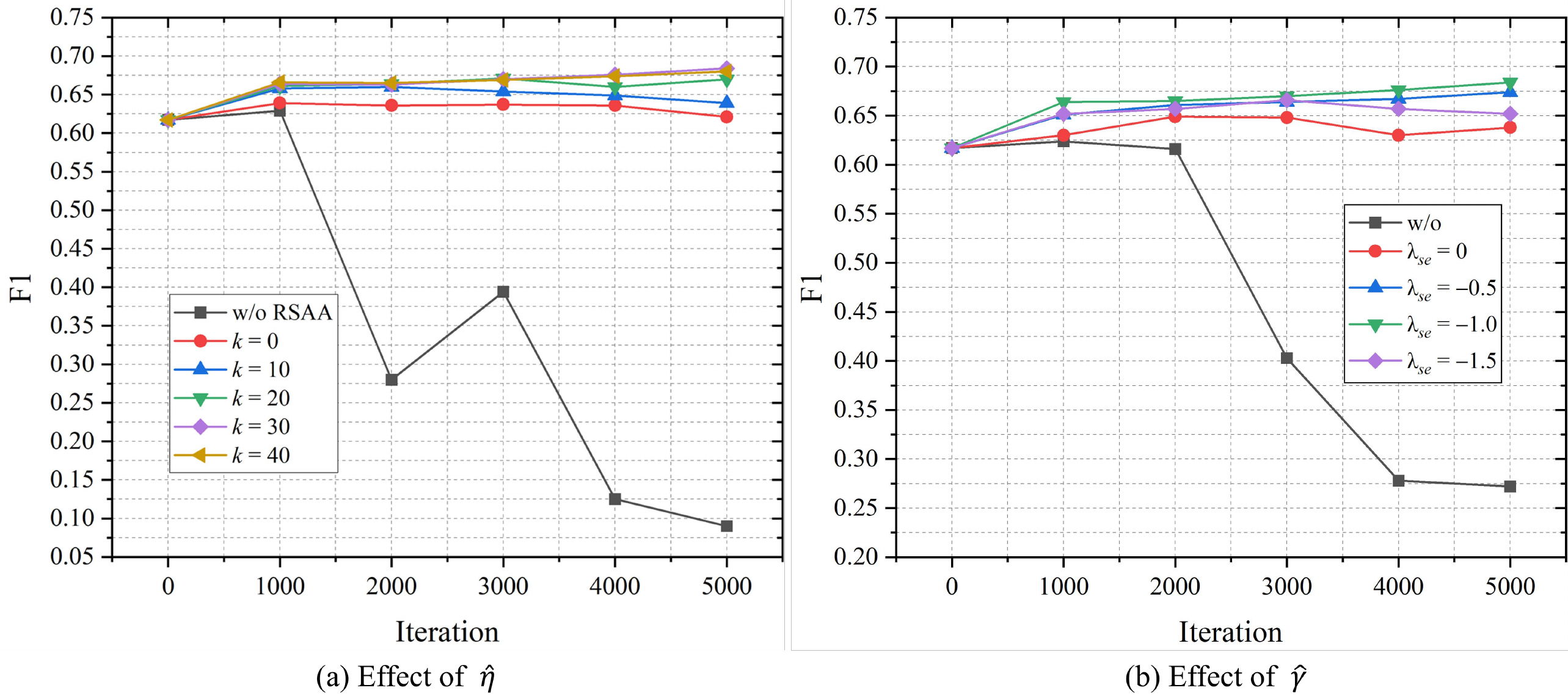}
        \caption{The benefits of \ouradja on performance during training iterations. (a) shows the effect of the \ourinsfactor $\hat \eta$ during the iterative process (with fixed $\lambdaSecmask=-1$), where `w/o \ouradja' in (a) represents the vanilla feature alignment using the L1 distance from Eq.\ref{eqrsaa}. (b) demonstrates the effect of the \ouradjafactor $\hat \gamma$ on performance during the iterative process (with fixed $k=30$); `w/o' in (b) indicates that the \ouradjafactor $\hat \gamma$ and secure set division are not utilized for feature alignment in Eq.\ref{eqrsaa}.}
        \label{fig:iterstep}
        \vspace{-5mm}
    \end{figure}

    \begin{figure}[h]
        \centering
        \includegraphics[width=0.9999\columnwidth]{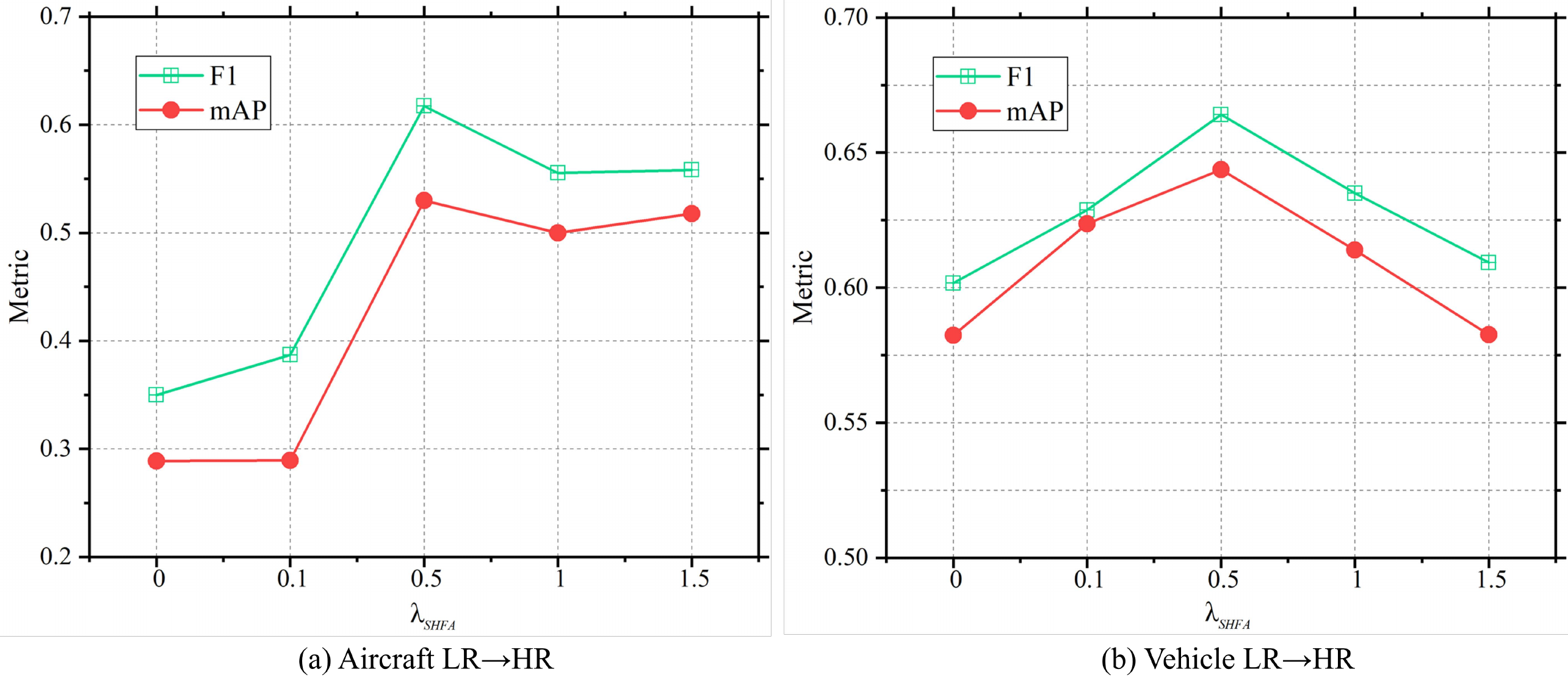}
        \caption{Hyperparameter discussions of $\lambdaSHFA$.}
        \label{fig:lambdashfa}
    \end{figure}

    \begin{figure}[h]
        \centering
        \includegraphics[width=0.999\columnwidth]{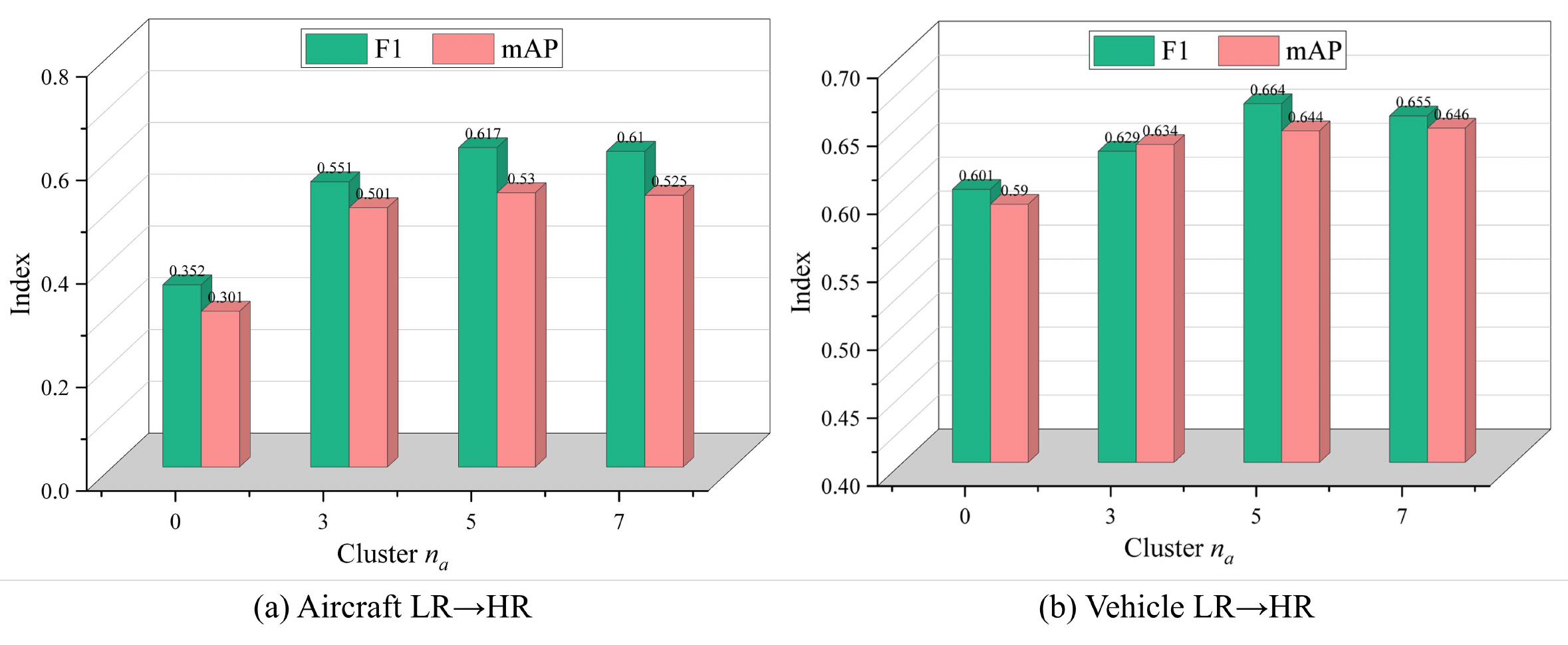}
        \caption{Hyperparameter discussions of cluster number $n_a$. When $n_a=0$, \ourda module degenerates into a conventional single-adversarial domain adaptation module.}
        \label{fig:na}
    \end{figure}

    \begin{figure}[h]
        \centering
        \includegraphics[width=0.9999\columnwidth]{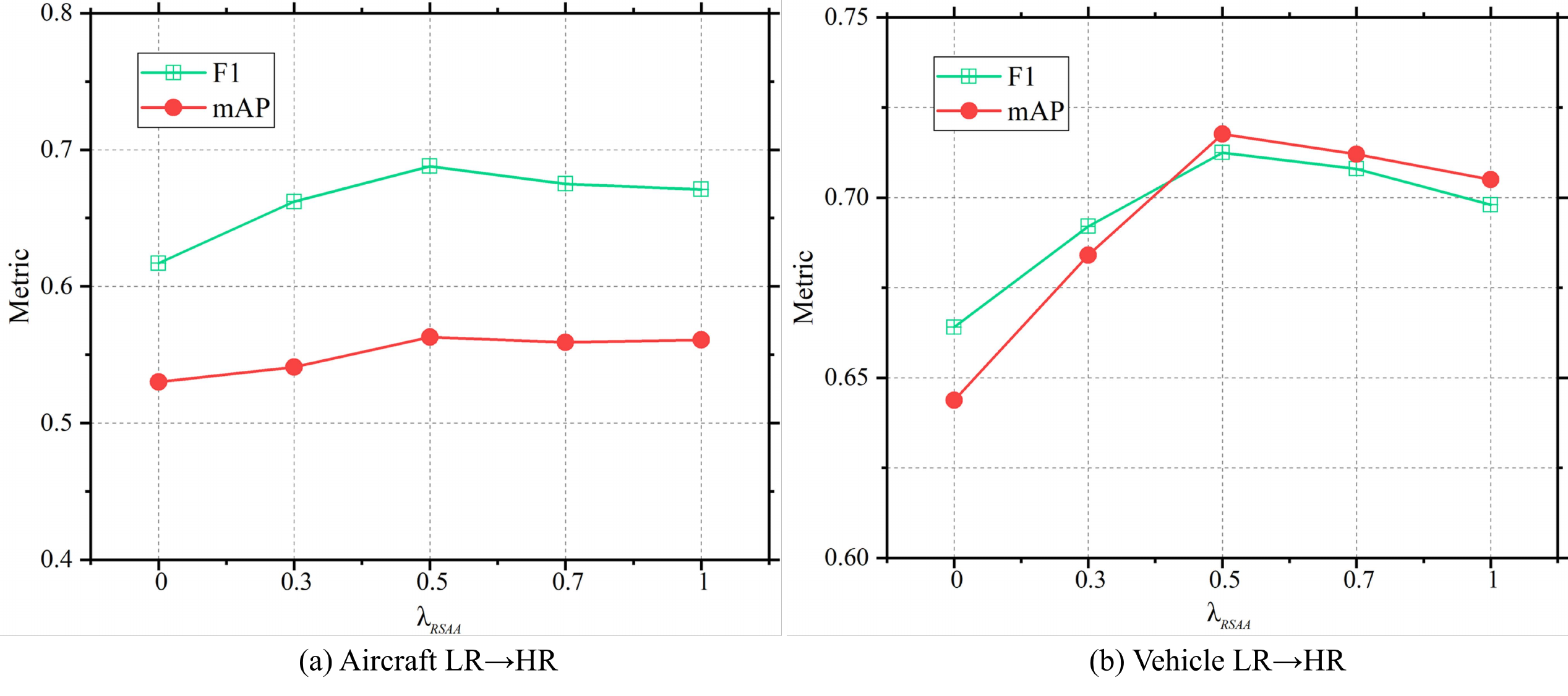}
        \caption{Hyperparameter discussions of $\lambdaRSAA$.}
        \label{fig:lambdarsaa}
    \end{figure}

    \begin{figure}[h]
        \centering
        \includegraphics[width=0.999\columnwidth]{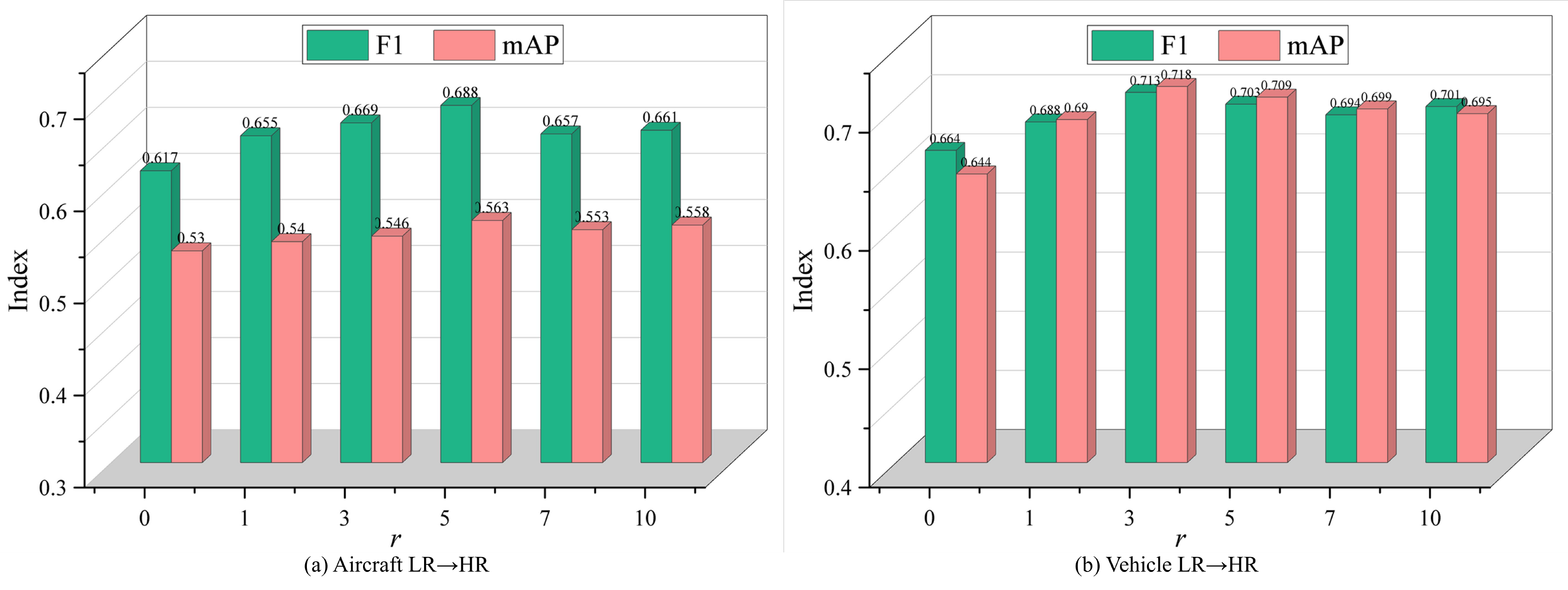}
        \caption{Hyperparameter discussions of $r$. $r=0$ means \ouradja module is not used during the training process.}
        \label{fig:r}
    \end{figure}

    \begin{figure}[h]
        \centering
        \includegraphics[width=0.6\columnwidth]{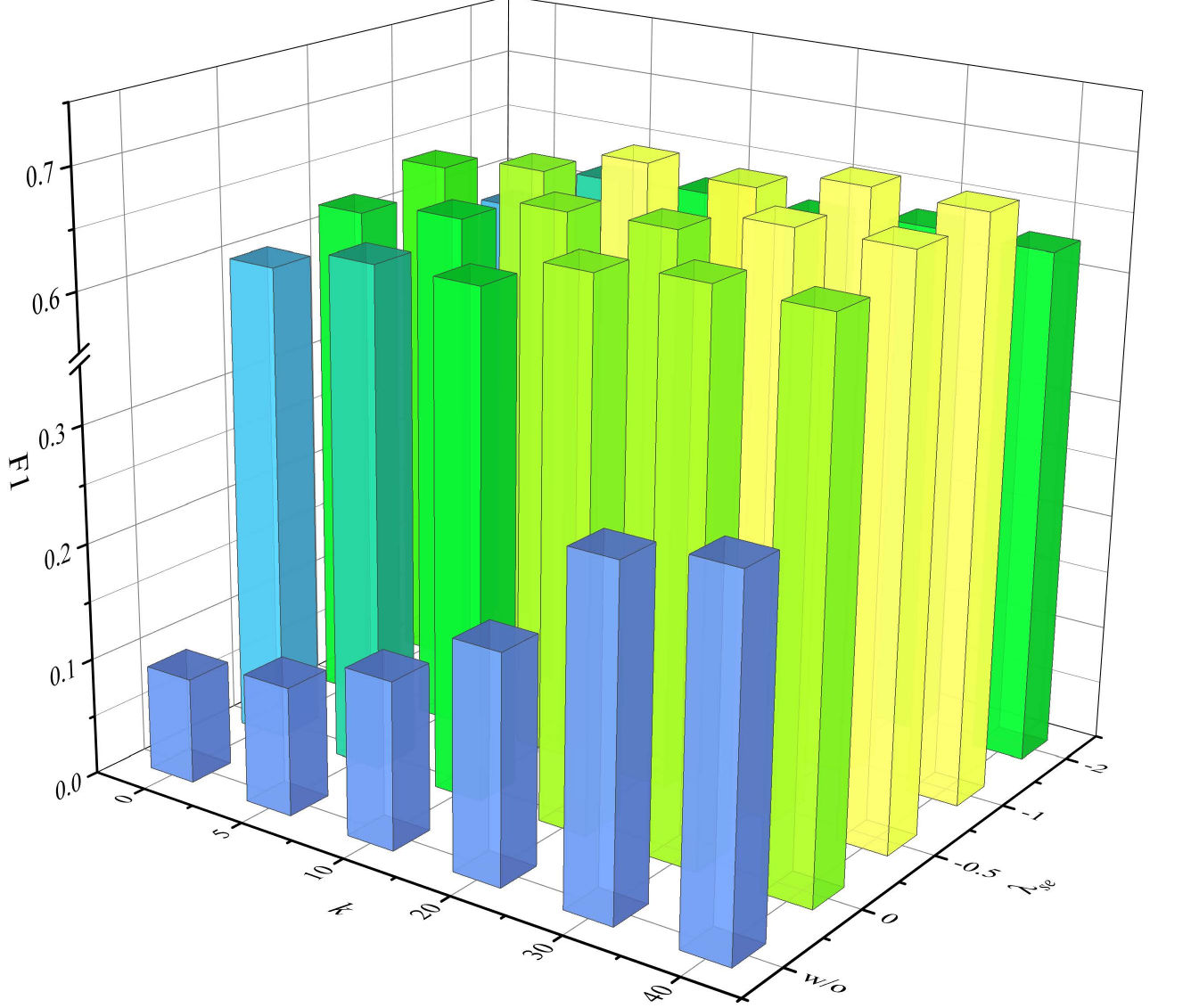}
        \caption{Hyperparameter discussions of $k$ and 
 $\lambdaSecmask$.}
        \label{fig:kmask}
    \end{figure}

    \subsection{Ablation Studies}
    The ablation studies are shown in Table~\ref{tab:ablation}. It is observed that, without introducing the \ourda and \ouradja module, the baseline Faster RCNN demonstrates limited cross-resolution detection performance. This indicates that, due to resolution differences, the original target detection model fails to achieve performance generalization for cross-resolution target detection.
    
    After introducing the \ourda module, the F1 score increases from 0.214 to 0.617 in the Aircraft LR→HR task). This demonstrates that the \ourda module effectively reduces domain discrepancies through structure-aware feature adaptation, thereby improving cross-resolution detection performance. 
    Furthermore, with the incorporation of the \ouradja module into the detector, the F1 score increases from 0.617 to 0.688, and the Precision rises from 0.648 to 0.722 in the Aircraft LR→HR task. The observed improvement in the Precision metric confirms that the \ouradja module enhances the model's discriminability in the target domain, which is also verified by the results of the Vehicle LR→HR task. The integration of either \ourda or \ouradja individually into the detection model leads to performance improvements, and the joint integration of both modules further achieves the most substantial benefits in cross-resolution detection performance.

    \begin{table}[]
        \centering
        \caption{Ablation studies on different components of \ourmethod. The bold figure denotes the best score.}
        \label{tab:ablation}
        \resizebox{\columnwidth}{!}{
        \begin{tabular}{@{}c|c|cccc@{}}
        \toprule
        Task                             & Ablations       & Recall & Precision & mAP   & F1    \\ \midrule
        \multirow{4}{*}{Aircraft LR→HR}  & Faster RCNN     & 0.172  & 0.284     & 0.146  & 0.214 \\
                                         & + \ourda        & 0.589  & 0.648     & 0.530  & 0.617 \\
                                         & + \ouradja      & 0.411  & 0.401     & 0.332  & 0.406 \\
                                         & + \ourda \& + \ouradja & \textbf{0.658}  & \textbf{0.722}     & \textbf{0.563}  & \textbf{0.688} \\ \midrule
        \multirow{4}{*}{Vehicle   LR→HR} & Faster RCNN     & 0.574  & 0.583     & 0.449  & 0.578 \\
                                         & + \ourda        & 0.644  & 0.686     & 0.644  & 0.664 \\
                                         & + \ouradja      & 0.611  & 0.660     & 0.576 & 0.634 \\
                                         & + \ourda \& + \ouradja & \textbf{0.723 } & \textbf{0.676}   & \textbf{0.718}  & \textbf{0.713} \\ \bottomrule
        \end{tabular}
        }
    \end{table}

    \subsection{Effective Analysis}
    \subsubsection{Effective Analysis of \ourda Module}
    Fig.\ref{fig:tsne} illustrates the feature adaption effects of the \ourda module on the Aircraft LR→HR and Vehicle LR→HR tasks. 
    As shown in Fig.\ref{fig:tsne} (a) and (c), before adaptation, the feature distributions of the source domain (orange) and the target domain (green) exhibit clear separation, with instances from the two domains forming distinct clusters. 
    This separation reflects the presence of distribution discrepancies and intra-domain diversity, which limit the performance of cross-resolution detection.
    In contrast, Fig.\ref{fig:tsne} (b) and (d) demonstrate that, after processing with the \ourda module, the feature distributions of the source and target domains show significant integration, with the distribution gap between them greatly reduced. Notably, in Fig.\ref{fig:tsne} (d) for the Vehicle dataset, the source and target domain features almost completely overlap, reflecting a high degree of distribution alignment. This indicates that the \ourda module exhibits strong alignment capability even in tasks with complex or highly divergent distributions.

    \subsubsection{Effective Analysis of \ouradja Module}
    According to the results shown in Fig.~\ref{fig:iterstep} (a), when the \ouradja module is not used and features are directly aligned using L1 loss, the F1 score drops sharply after 1000 training iterations. 
    When the \ourinsfactor $\hat \eta$ is removed (setting $k=0$), the F1 score initially increases but then decreases as training iterations progress. 
    In contrast, when larger values ($k=30$ or $k=40$) are set, the F1 score steadily increases during training and shows significant improvement compared to the case without $\hat \eta$. This indicates that the \ourinsfactor plays a positive role in mitigating error accumulation during the adjacency alignment process. Fig.~\ref{fig:iterstep} (b) illustrates the impact of the \ouradjafactor for iterative adjacency semantic alignment process. 
    According to the definition of $\hat \gamma$ in Eq.~\ref{eq:securemaskmu}, the smaller $\lambdaSecmask$ indicates the stronger the filtering effect on the risky set. Based on the results in Fig.~\ref{fig:iterstep}, when $\hat \gamma$ is removed, the model experiences a sharp collapse in the F1 score after approximately 2000 iterations during the adjacency semantic alignment process. 
    In contrast, when $\hat \gamma$ is introduced, the F1 score shows a gradual upward trend throughout the iterations. This demonstrates that the \ouradjafactor $\hat \gamma$ helps to filter out unreliable risky sets, preventing the propagation of risky information, which is critical for reliable adjacency alignment. 
    To acquire satisfactory  performance, the hyperparameters are set to $k=30$ and $\lambdaSecmask=-1$ by default, as indicated by the results in Fig.~\ref{fig:iterstep}.

    \subsection{Hyperparameter Discussions}
    \label{discuss}
    
    \subsubsection{Discussions on Hyperparameter of \ourda}
    Fig.~\ref{fig:lambdashfa} illustrates the effect of the hyperparameter $\lambdaSHFA$ from Eq.~\ref{eq:loss}, on cross-resolution target detection performance ($\lambdaRSAA=0$, $n_a=5$). 
    The results show that both F1 score and mAP initially increase and then decrease as $\lambdaSHFA$ increases, indicating that appropriate tuning of $\lambdaSHFA$ is crucial for optimal model performance. 
    For the Aircraft LR→HR and Vehicle LR→HR tasks, the best results are achieved when $\lambdaSHFA=0.5$, where both F1 and mAP reach their highest values.
    When $\lambdaSHFA$ is small (e.g., 0 or 0.1), the model exhibits insufficient feature adaptation capability for cross-resolution detection, resulting in lower performance. As $\lambdaSHFA$ increases to 0.5, there are significant improvements in F1 and mAP, demonstrating the effectiveness of the \ourda module. 
    However, when $\lambdaSHFA$ becomes too large (e.g., 1.0 or 1.5), excessive alignment constraints may impair the model’s generalization ability to the target domain, leading to a performance drop.
    From the Vehicle LR→HR results in Fig.~\ref{fig:lambdashfa}, the F1 and mAP of the Vehicle dataset consistently outperform those of the Aircraft dataset. This is because the Vehicle dataset exhibits a more consistent target scale distribution and more similar imaging sensor, which makes feature adaptation more feasible. According to the aforementioned findings, it is recommended to set $\lambdaSHFA$ to around 0.5 to achieve a balance between feature adaptation and generalization capability.

    {Fig.~\ref{fig:na} illustrates the effect of the number of structure anchors $n_a$ (defined in Section~\ref{SECSHFA}), with $\lambdaSHFA=0.5$. 
     Each structure anchor corresponds to one domain discriminator.
    Ideally, $n_a$ should match the actual number of typical structural types in the source domain, ensuring that each type corresponds to a unique domain discriminator for optimal hierarchical feature adaptation. 
    Setting $n_a$ too small may group features from different structural types together, causing confusion and suboptimal adaptation. Conversely, setting $n_a$ too large may assign multiple discriminators to the same structural type, resulting in competition effects and reduced adaptation effectiveness.
    For both the Aircraft LR→HR and Vehicle LR→HR tasks, the results in Fig~\ref{fig:na} show that the F1 and mAP score vary with different $n_a$. 
    When $n_a$ is set to 0, the \ourda module degenerates into a conventional single-adversarial domain adaptation model, resulting in the lowest F1 and mAP values. 
    As $n_a$ increases, introducing more hierarchical discriminators into the \ourda module, both F1 and mAP improve, indicating that assigning features to discriminators based on structural similarity enhances feature adaptation performance. 
    The best results are observed when $n_a=5$, suggesting that the number of domain discriminators matches the typical structural types in the source domain, achieving optimal hierarchical feature adaptation. Further increasing $n_a$ introduces excessive domain discriminators, leading to competition effects and a slight decline in performance. Therefore, it is recommended to set $n_a$ to around 5 to ensure optimal performance.}

    \subsubsection{Discussions on Hyperparameter of \ouradja} 
    Fig.~\ref{fig:lambdarsaa} shows the impact of hyperparameter $\lambdaRSAA$ (defined in Eq.~\ref{eq:loss}) on cross-resolution target detection performance. 
    In both the Aircraft LR→HR and Vehicle LR→HR tasks, as $\lambdaRSAA$ increases from 0 to 0.5, F1 and mAP gradually reach their peak values. These indicate that, under an appropriate adjacency semantic alignment strength, \ouradja can effectively enhance the detector's discriminative capability in the target domain. 
    When $\lambdaRSAA>0.5$, the performance tends to level off or even decline, suggesting that further increasing $\lambdaRSAA$ provides limited benefits, and the excessive alignment may even restrict the generalization ability. 
    Therefore, setting $\lambdaRSAA$ to around 0.5 can ensure satisfactory cross-resolution detection performance.

    Fig.~\ref{fig:r} illustrates the impact of the hyperparameter $r$ on detection performance. 
    {While a larger $r$ enables improved semantic alignment by capturing richer neighbor semantic information, it also increases the risk of introducing noise-instances and error accumulation.
    For the Aircraft LR→HR task, both F1 and mAP steadily improve as $r$ increases from 0 to 5, reaching their highest values at $r=5$. In contrast, for the Vehicle LR→HR task, optimal performance is achieved at $r=3$. Notably, further increasing $r$ beyond these points leads to performance declines in both datasets.
    This phenomenon occurs because an excessively large $r$ may introduce semantically inconsistent noise-instances, leading to error accumulation during semantic alignment.
    Moreover, the optimal $r$ for the Vehicle dataset is lower than that for the Aircraft dataset. The difference can be attributed to the fact that vehicle targets typically appear in urban scenes with more complex and diverse contexts, which increases the proportion of structurally similar but semantically different noise-instances (e.g., containers, air conditioner units). Based on the above analysis, the optimal value of $r$ varies across different datasets and should be carefully selected to maximize detection performance. In this paper, we set $r=5$ for the Aircraft dataset and $r=3$ for the Vehicle dataset by default.}

    Fig.~\ref{fig:kmask} illustrates the impact of the hyper-parameters $k$ and $\lambdaSecmask$ on cross-resolution detection performance. 
    $k$ acts as a scaling factor for the \ourinsfactor $\hat \eta$ in Eq.~\ref{eq:k}, directly influencing the strength of adjacency alignment. 
    Meanwhile, $\lambdaSecmask$ determines the selection range of the secure adjacency set in Eq.~\ref{eq:securemaskmu}. 
    As shown in Fig.~\ref{fig:kmask}, the F1 score reaches its maximum when $k=30$ and $\lambdaSecmask=-1$. When $k=0$, the model loses its ability to distinguish between reliable and unreliable instances. As a result, the adjacency alignment process treats all instances equally, leading to the accumulation of errors. 
    An excessively large $\lambdaSecmask$ causes the secure set to include more risky information, which leads to error propagation and weakens the model's discriminative capability. 
    Conversely, a small $\lambdaSecmask$ filters out too many informative instances, thereby degrading the adjacency alignment performance.

    \section{Conclusion}
    \label{conclusion}
    
    In this paper, we present a novel SAR target detection framework (\ourmethod) that effectively addresses the challenges of cross-resolution SAR detection. 
    \ourmethod achieves reliable cross-resolution detection by incorporating evidential learning theory and target scattering structure correlations, thereby evolving into two key modules: Structure-induced Hierarchical Feature Adaptation (\ourda) and Reliable Structural Adjacency Alignment (\ouradja).
    Specifically, By exploring multi-mode structures within the domain, \ourmethod integrates \ourda to perform fine-grained distribution adaptation under the guidance of structure similarity, thereby alleviating feature misalignment and enhancing the interpretability of the adaptation process.
    Furthermore, \ouradja is integrated to improve the discriminability and generalization ability  of \ourmethod by transferring the discriminative semantics from source reliable adjacency samples.
    This dual approach enables \ourmethod to significantly promote positive transfer while avoiding negative transfer.
    Extensive experimental results demonstrate that \ourmethod  improves the performance of  cross-resolution SAR target detection.
    In the future, we will continue to explore how to leverage foundation models to further enhance cross-resolution target detection performance, thereby promoting the practical application of this technology.

    \balance
    \bibliography{longstrings,ref}

\begin{thebibliography}{10}
\providecommand{\url}[1]{#1}
\csname url@samestyle\endcsname
\providecommand{\newblock}{\relax}
\providecommand{\bibinfo}[2]{#2}
\providecommand{\BIBentrySTDinterwordspacing}{\spaceskip=0pt\relax}
\providecommand{\BIBentryALTinterwordstretchfactor}{4}
\providecommand{\BIBentryALTinterwordspacing}{\spaceskip=\fontdimen2\font plus
\BIBentryALTinterwordstretchfactor\fontdimen3\font minus \fontdimen4\font\relax}
\providecommand{\BIBforeignlanguage}[2]{{%
\expandafter\ifx\csname l@#1\endcsname\relax
\typeout{** WARNING: IEEEtran.bst: No hyphenation pattern has been}%
\typeout{** loaded for the language `#1'. Using the pattern for}%
\typeout{** the default language instead.}%
\else
\language=\csname l@#1\endcsname
\fi
#2}}
\providecommand{\BIBdecl}{\relax}
\BIBdecl

\bibitem{sartutorials}
A.~Moreira, P.~Prats-Iraola, M.~Younis, G.~Krieger, I.~Hajnsek, and K.~P. Papathanassiou, ``{A tutorial on synthetic aperture radar},'' \emph{IEEE Geoscience and Remote Sensing Magazine}, vol.~1, no.~1, pp. 6--43, 2013.

\bibitem{Liu04072025}
\BIBentryALTinterwordspacing
C.~Liu, Z.~Zhang, M.~Wang, S.~Xiang, and G.~Xie, ``{A novel cross fusion model with fine-grained detail reconstruction for remote sensing image pan-sharpening},'' \emph{Geo-spatial Information Science}, vol.~28, no.~4, pp. 1520--1548, 2025. [Online]. Available: \url{https://doi.org/10.1080/10095020.2024.2416899}
\BIBentrySTDinterwordspacing

\bibitem{zhang2023frequency}
L.~Zhang, Y.~Liu, W.~Zhao, X.~Wang, G.~Li, and Y.~He, ``{Frequency-adaptive learning for SAR ship detection in clutter scenes},'' \emph{IEEE Transactions on Geoscience and Remote Sensing}, vol.~61, pp. 1--14, 2023.

\bibitem{AIRcraft-1.0}
W.~Zhirui, K.~Yuzhuo, Z.~Xuan, W.~Yuelei, Z.~Ting, and S.~Xian, ``{SAR-AIRcraft-1.0: High-resolution SAR aircraft detection and recognition dataset},'' \emph{Journal of Radars}, vol.~12, no.~4, pp. 906--922, 2023.

\bibitem{DAN}
M.~Long, Y.~Cao, J.~Wang, and M.~Jordan, ``{Learning transferable features with deep adaptation networks},'' in \emph{International conference on machine learning}.\hskip 1em plus 0.5em minus 0.4em\relax PMLR, 2015, pp. 97--105.

\bibitem{DANN}
Y.~Ganin and V.~Lempitsky, ``{Unsupervised domain adaptation by backpropagation},'' in \emph{International conference on machine learning}.\hskip 1em plus 0.5em minus 0.4em\relax PMLR, 2015, pp. 1180--1189.

\bibitem{ida}
B.~Pan, Z.~Xu, T.~Shi, T.~Li, and Z.~Shi, ``{An imbalanced discriminant alignment approach for domain adaptive SAR ship detection},'' \emph{IEEE Transactions on Geoscience and Remote Sensing}, vol.~61, pp. 1--11, 2023.

\bibitem{HSA}
J.~Zhang, S.~Li, Y.~Dong, B.~Pan, and Z.~Shi, ``{Hierarchical similarity alignment for domain adaptive ship detection in SAR images},'' \emph{IEEE Transactions on Geoscience and Remote Sensing}, vol.~60, pp. 1--11, 2022.

\bibitem{huang2024domain}
H.~Huang, J.~Guo, H.~Lin, Y.~Huang, and X.~Ding, ``{Domain Adaptive Oriented Object Detection from Optical to SAR Images},'' \emph{IEEE Transactions on Geoscience and Remote Sensing}, 2024.

\bibitem{zhang2025cross}
X.~Zhang, S.~Zhang, Z.~Sun, C.~Liu, Y.~Sun, K.~Ji, and G.~Kuang, ``{Cross-Sensor SAR Image Target Detection Based on Dynamic Feature Discrimination and Center-Aware Calibration},'' \emph{IEEE Transactions on Geoscience and Remote Sensing}, 2025.

\bibitem{shi2021unsupervised}
Y.~Shi, L.~Du, and Y.~Guo, ``{Unsupervised domain adaptation for SAR target detection},'' \emph{IEEE Journal of Selected Topics in Applied Earth Observations and Remote Sensing}, vol.~14, pp. 6372--6385, 2021.

\bibitem{zou2023cross}
B.~Zou, J.~Qin, and L.~Zhang, ``{Cross-scene target detection based on feature adaptation and uncertainty-aware pseudo-label learning for high resolution SAR images},'' \emph{ISPRS Journal of Photogrammetry and Remote Sensing}, vol. 200, pp. 173--190, 2023.

\bibitem{zhao2022automatic}
S.~Zhao, Z.~Zhang, W.~Guo, and Y.~Luo, ``{An automatic ship detection method adapting to different satellites SAR images with feature alignment and compensation loss},'' \emph{IEEE Transactions on Geoscience and Remote Sensing}, vol.~60, pp. 1--17, 2022.

\bibitem{zhao2022feature}
S.~Zhao, Y.~Luo, T.~Zhang, W.~Guo, and Z.~Zhang, ``{A feature decomposition-based method for automatic ship detection crossing different satellite SAR images},'' \emph{IEEE Transactions on Geoscience and Remote Sensing}, vol.~60, pp. 1--15, 2022.

\bibitem{HTCN}
C.~Chen, Z.~Zheng, X.~Ding, Y.~Huang, and Q.~Dou, ``{Harmonizing Transferability and Discriminability for Adapting Object Detectors},'' in \emph{2020 IEEE/CVF Conference on Computer Vision and Pattern Recognition (CVPR)}, 2020, pp. 8866--8875.

\bibitem{kim2019self}
S.~Kim, J.~Choi, T.~Kim, and C.~Kim, ``{Self-training and adversarial background regularization for unsupervised domain adaptive one-stage object detection},'' in \emph{Proceedings of the IEEE/CVF international conference on computer vision}, 2019, pp. 6092--6101.

\bibitem{wang2018sar}
Z.~Wang, L.~Du, J.~Mao, B.~Liu, and D.~Yang, ``{SAR target detection based on SSD with data augmentation and transfer learning},'' \emph{IEEE Geoscience and Remote Sensing Letters}, vol.~16, no.~1, pp. 150--154, 2018.

\bibitem{SCEDet}
B.~Zou, J.~Qin, and L.~Zhang, ``{Vehicle Detection Based on Semantic-Context Enhancement for High-Resolution SAR Images in Complex Background},'' \emph{IEEE Geoscience and Remote Sensing Letters}, vol.~19, pp. 1--5, 2022.

\bibitem{10934049}
C.~Liu, Z.~Zhang, and M.~Wang, ``{RAMSF: A Novel Generic Framework for Optical Remote Sensing Multimodal Spatial-Spectral Fusion},'' \emph{IEEE Transactions on Geoscience and Remote Sensing}, vol.~63, pp. 1--22, 2025.

\bibitem{sensoy2018evidential}
M.~Sensoy, L.~Kaplan, and M.~Kandemir, ``{Evidential deep learning to quantify classification uncertainty},'' \emph{Advances in neural information processing systems}, vol.~31, 2018.

\bibitem{pei2024evidential}
J.~Pei, A.~Men, Y.~Liu, X.~Zhuang, and Q.~Chen, ``{Evidential multi-source-free unsupervised domain adaptation},'' \emph{IEEE Transactions on Pattern Analysis and Machine Intelligence}, vol.~46, no.~8, pp. 5288--5305, 2024.

\bibitem{maddox2019simple}
W.~J. Maddox, P.~Izmailov, T.~Garipov, D.~P. Vetrov, and A.~G. Wilson, ``{A simple baseline for bayesian uncertainty in deep learning},'' \emph{Advances in neural information processing systems}, vol.~32, 2019.

\bibitem{gal2016dropout}
Y.~Gal and Z.~Ghahramani, ``{Dropout as a bayesian approximation: Representing model uncertainty in deep learning},'' in \emph{international conference on machine learning}.\hskip 1em plus 0.5em minus 0.4em\relax PMLR, 2016, pp. 1050--1059.

\bibitem{lakshminarayanan2017simple}
B.~Lakshminarayanan, A.~Pritzel, and C.~Blundell, ``{Simple and scalable predictive uncertainty estimation using deep ensembles},'' \emph{Advances in neural information processing systems}, vol.~30, 2017.

\bibitem{zhang2023provable}
Q.~Zhang, H.~Wu, C.~Zhang, Q.~Hu, H.~Fu, J.~T. Zhou, and X.~Peng, ``{Provable dynamic fusion for low-quality multimodal data},'' in \emph{International conference on machine learning}.\hskip 1em plus 0.5em minus 0.4em\relax PMLR, 2023, pp. 41\,753--41\,769.

\bibitem{10656006}
J.~Chen, B.~Ma, H.~Cui, and Y.~Xia, ``{Think Twice Before Selection: Federated Evidential Active Learning for Medical Image Analysis with Domain Shifts},'' in \emph{2024 IEEE/CVF Conference on Computer Vision and Pattern Recognition (CVPR)}, 2024, pp. 11\,439--11\,449.

\bibitem{10948323}
C.~Huang, W.~Huang, Q.~Jiang, W.~Wang, J.~Wen, and B.~Zhang, ``{Multimodal Evidential Learning for Open-World Weakly-Supervised Video Anomaly Detection},'' \emph{IEEE Transactions on Multimedia}, pp. 1--12, 2025.

\bibitem{zhou2024outlier}
Y.~Zhou, H.~Xia, D.~Yu, J.~Cheng, and J.~Li, ``{Outlier detection method based on high-density iteration},'' \emph{Information Sciences}, vol. 662, p. 120286, 2024.

\bibitem{focal}
T.-Y. Lin, P.~Goyal, R.~Girshick, K.~He, and P.~Doll{\'a}r, ``Focal loss for dense object detection,'' in \emph{Proceedings of the IEEE international conference on computer vision}, 2017, pp. 2980--2988.

\bibitem{yolov7}
C.-Y. Wang, A.~Bochkovskiy, and H.-Y.~M. Liao, ``Yolov7: Trainable bag-of-freebies sets new state-of-the-art for real-time object detectors,'' in \emph{Proceedings of the IEEE/CVF conference on computer vision and pattern recognition}, 2023, pp. 7464--7475.

\bibitem{fasterrcnn}
S.~Ren, K.~He, R.~Girshick, and J.~Sun, ``{Faster R-CNN: Towards real-time object detection with region proposal networks},'' \emph{Advances in neural information processing systems}, vol.~28, 2015.

\bibitem{sparsercnn}
P.~Sun, R.~Zhang, Y.~Jiang, T.~Kong, C.~Xu, W.~Zhan, M.~Tomizuka, L.~Li, Z.~Yuan, C.~Wang \emph{et~al.}, ``{Sparse R-CNN: End-to-end object detection with learnable proposals},'' in \emph{Proceedings of the IEEE/CVF conference on computer vision and pattern recognition}, 2021, pp. 14\,454--14\,463.

\bibitem{10091564}
Y.~Zhou, H.~Liu, F.~Ma, Z.~Pan, and F.~Zhang, ``{A Sidelobe-Aware Small Ship Detection Network for Synthetic Aperture Radar Imagery},'' \emph{IEEE Transactions on Geoscience and Remote Sensing}, vol.~61, pp. 1--16, 2023.

\bibitem{10285445}
M.~Ju, B.~Niu, and J.~Zhang, ``{FPDDet: An Efficient Rotated SAR Ship Detector Based on Simple Polar Encoding and Decoding},'' \emph{IEEE Transactions on Geoscience and Remote Sensing}, vol.~61, pp. 1--15, 2023.

\bibitem{10056331}
D.~Zhao, Z.~Chen, Y.~Gao, and Z.~Shi, ``{Classification Matters More: Global Instance Contrast for Fine-Grained SAR Aircraft Detection},'' \emph{IEEE Transactions on Geoscience and Remote Sensing}, vol.~61, pp. 1--15, 2023.

\bibitem{qin2024scattering}
J.~Qin, B.~Zou, Y.~Chen, H.~Li, and L.~Zhang, ``{Scattering Attribute Embedded Network for Few-Shot SAR ATR},'' \emph{IEEE Transactions on Aerospace and Electronic Systems}, 2024.

\bibitem{10813601}
Y.~Yang, Y.~Du, L.~Zhang, G.~Li, Y.~Chen, G.~Cheng, and S.~Song, ``{DAFDet: A Unified Dynamic SAR Target Detection Architecture With Asymptotic Fusion Enhancement and Feature Encoding Decoupling},'' \emph{IEEE Transactions on Geoscience and Remote Sensing}, vol.~63, pp. 1--22, 2025.

\bibitem{8763918}
Z.~Cui, Q.~Li, Z.~Cao, and N.~Liu, ``{Dense Attention Pyramid Networks for Multi-Scale Ship Detection in SAR Images},'' \emph{IEEE Transactions on Geoscience and Remote Sensing}, vol.~57, no.~11, pp. 8983--8997, 2019.

\bibitem{10294268}
C.~Liu, L.~Wei, Z.~Zhang, X.~Feng, and S.~Xiang, ``{Recursive Self-Attention Modules-Based Network for Panchromatic and Multispectral Image Fusion},'' \emph{IEEE Journal of Selected Topics in Applied Earth Observations and Remote Sensing}, vol.~16, pp. 10\,067--10\,083, 2023.

\bibitem{cheng2022inshore}
J.~Cheng, D.~Xiang, J.~Tang, Y.~Zheng, D.~Guan, and B.~Du, ``{Inshore ship detection in large-scale SAR images based on saliency enhancement and Bhattacharyya-like distance},'' \emph{Remote Sensing}, vol.~14, no.~12, p. 2832, 2022.

\bibitem{ren2025confucius}
P.~Ren, Z.~Han, Z.~Yu, and B.~Zhang, ``{Confucius tri-learning: A paradigm of learning from both good examples and bad examples},'' \emph{Pattern Recognition}, vol. 163, p. 111481, 2025.

\bibitem{9916304}
M.~Ju and Q.~Hu, ``{Vision-Inspired Filtering Algorithm for SAR Ship Detection Based on Generative Adversarial Networks},'' \emph{IEEE Geoscience and Remote Sensing Letters}, vol.~19, pp. 1--5, 2022.

\bibitem{goodfellow2014generative}
I.~J. Goodfellow, J.~Pouget-Abadie, M.~Mirza, B.~Xu, D.~Warde-Farley, S.~Ozair, A.~Courville, and Y.~Bengio, ``{Generative adversarial nets},'' \emph{Advances in neural information processing systems}, vol.~27, 2014.

\bibitem{8950292}
Z.~Ren, B.~Hou, Q.~Wu, Z.~Wen, and L.~Jiao, ``{A Distribution and Structure Match Generative Adversarial Network for SAR Image Classification},'' \emph{IEEE Transactions on Geoscience and Remote Sensing}, vol.~58, no.~6, pp. 3864--3880, 2020.

\bibitem{10054495}
L.~Zhang, Y.~Liu, W.~Zhao, X.~Wang, G.~Li, and Y.~He, ``{Frequency-Adaptive Learning for SAR Ship Detection in Clutter Scenes},'' \emph{IEEE Transactions on Geoscience and Remote Sensing}, vol.~61, pp. 1--14, 2023.

\bibitem{10770564}
Y.~Dai, M.~Zou, Y.~Li, X.~Li, K.~Ni, and J.~Yang, ``{DenoDet: Attention as Deformable Multisubspace Feature Denoising for Target Detection in SAR Images},'' \emph{IEEE Transactions on Aerospace and Electronic Systems}, vol.~61, no.~2, pp. 4729--4743, 2025.

\bibitem{10623223}
Y.~Suo, Y.~Wu, T.~Miao, W.~Diao, X.~Sun, and K.~Fu, ``{Adaptive SAR Image Enhancement for Aircraft Detection via Speckle Suppression and Channel Combination},'' \emph{IEEE Transactions on Geoscience and Remote Sensing}, vol.~62, pp. 1--15, 2024.

\bibitem{MADA}
Z.~Pei, Z.~Cao, M.~Long, and J.~Wang, ``{Multi-adversarial domain adaptation},'' in \emph{Proceedings of the AAAI conference on artificial intelligence}, vol.~32, no.~1, 2018.

\bibitem{khodabandeh2019robust}
M.~Khodabandeh, A.~Vahdat, M.~Ranjbar, and W.~G. Macready, ``{A robust learning approach to domain adaptive object detection},'' in \emph{Proceedings of the IEEE/CVF international conference on computer vision}, 2019, pp. 480--490.

\bibitem{qin2024conditional}
J.~Qin, K.~Wang, B.~Zou, L.~Zhang, and J.~van~de Weijer, ``{Conditional Diffusion Model with Spatial-Frequency Refinement for SAR-to-Optical Image Translation},'' \emph{IEEE Transactions on Geoscience and Remote Sensing}, 2024.

\bibitem{deng2021unbiased}
J.~Deng, W.~Li, Y.~Chen, and L.~Duan, ``{Unbiased mean teacher for cross-domain object detection},'' in \emph{Proceedings of the IEEE/CVF conference on computer vision and pattern recognition}, 2021, pp. 4091--4101.

\bibitem{cat}
M.~Kennerley, J.-G. Wang, B.~Veeravalli, and R.~T. Tan, ``{CAT: Exploiting inter-class dynamics for domain adaptive object detection},'' in \emph{Proceedings of the IEEE/CVF Conference on Computer Vision and Pattern Recognition}, 2024, pp. 16\,541--16\,550.

\bibitem{deng2023harmonious}
J.~Deng, D.~Xu, W.~Li, and L.~Duan, ``{Harmonious teacher for cross-domain object detection},'' in \emph{Proceedings of the IEEE/CVF conference on computer vision and pattern recognition}, 2023, pp. 23\,829--23\,838.

\bibitem{lu2025visual}
W.~Lu, J.~Wang, T.~Wang, K.~Zhang, X.~Jiang, and H.~Zhao, ``{Visual style prompt learning using diffusion models for blind face restoration},'' \emph{Pattern Recognition}, vol. 161, p. 111312, 2025.

\bibitem{sadis}
J.~Qin, S.~Li, A.~Gomez-Villa, S.~Yang, Y.~Wang, K.~Wang, and J.~van~de Weijer, ``Free-lunch color-texture disentanglement for stylized image generation,'' \emph{arXiv preprint arXiv:2503.14275}, 2025.

\bibitem{8995481}
L.~Zhang, X.~Wang, D.~Yang, T.~Sanford, S.~Harmon, B.~Turkbey, B.~J. Wood, H.~Roth, A.~Myronenko, D.~Xu, and Z.~Xu, ``{Generalizing Deep Learning for Medical Image Segmentation to Unseen Domains via Deep Stacked Transformation},'' \emph{IEEE Transactions on Medical Imaging}, vol.~39, no.~7, pp. 2531--2540, 2020.

\bibitem{yue2021transporting}
Z.~Yue, Q.~Sun, X.-S. Hua, and H.~Zhang, ``{Transporting causal mechanisms for unsupervised domain adaptation},'' in \emph{Proceedings of the IEEE/CVF International Conference on Computer Vision}, 2021, pp. 8599--8608.

\bibitem{wang2025crowdvlm}
Z.~Wang, P.~Feng, Y.~Lin, S.~Cai, Z.~Bian, J.~Yan, and X.~Zhu, ``{Crowdvlm-r1: Expanding r1 ability to vision language model for crowd counting using fuzzy group relative policy reward},'' \emph{arXiv preprint arXiv:2504.03724}, 2025.

\bibitem{10767752}
J.~Qin, B.~Zou, H.~Li, and L.~Zhang, ``{Efficient End-to-End Diffusion Model for One-Step SAR-to-Optical Translation},'' \emph{IEEE Geoscience and Remote Sensing Letters}, vol.~22, pp. 1--5, 2025.

\bibitem{shi2022unsupervised}
Y.~Shi, L.~Du, Y.~Guo, and Y.~Du, ``{Unsupervised domain adaptation based on progressive transfer for ship detection: From optical to SAR images},'' \emph{IEEE Transactions on Geoscience and Remote Sensing}, vol.~60, pp. 1--17, 2022.

\bibitem{10570480}
S.~Liu, D.~Li, J.~Wan, C.~Zheng, J.~Su, H.~Liu, and H.~Zhu, ``{Source-Assisted Hierarchical Semantic Calibration Method for Ship Detection Across Different Satellite SAR Images},'' \emph{IEEE Transactions on Geoscience and Remote Sensing}, vol.~62, pp. 1--21, 2024.

\bibitem{cyclegan}
J.-Y. Zhu, T.~Park, P.~Isola, and A.~A. Efros, ``Unpaired image-to-image translation using cycle-consistent adversarial networks,'' in \emph{Proceedings of the IEEE international conference on computer vision}, 2017, pp. 2223--2232.

\bibitem{pu2023ship}
X.~Pu, H.~Jia, Y.~Xin, F.~Wang, and H.~Wang, ``{Ship detection in low-quality SAR images via an unsupervised domain adaption method},'' \emph{Remote Sensing}, vol.~15, no.~13, p. 3326, 2023.

\bibitem{emd}
Y.~Rubner, C.~Tomasi, and L.~J. Guibas, ``The earth mover's distance as a metric for image retrieval,'' \emph{International journal of computer vision}, vol.~40, pp. 99--121, 2000.

\bibitem{sinkhorn}
M.~Cuturi, ``Sinkhorn distances: Lightspeed computation of optimal transport,'' \emph{Advances in neural information processing systems}, vol.~26, 2013.

\bibitem{feng2014note}
P.~Feng and X.~Peng, ``{A note on Monge--Kantorovich problem},'' \emph{Statistics \& Probability Letters}, vol.~84, pp. 204--211, 2014.

\bibitem{tsne}
L.~Van~der Maaten and G.~Hinton, ``Visualizing data using t-sne.'' \emph{Journal of machine learning research}, vol.~9, no.~11, 2008.

\bibitem{daf}
Y.~Chen, W.~Li, C.~Sakaridis, D.~Dai, and L.~Van~Gool, ``{Domain adaptive Faster R-CNN for object detection in the wild},'' in \emph{Proceedings of the IEEE conference on computer vision and pattern recognition}, 2018, pp. 3339--3348.

\bibitem{zhang2022sefepnet}
P.~Zhang, H.~Xu, T.~Tian, P.~Gao, L.~Li, T.~Zhao, N.~Zhang, and J.~Tian, ``{SEFEPNet: Scale expansion and feature enhancement pyramid network for SAR aircraft detection with small sample dataset},'' \emph{IEEE Journal of Selected Topics in Applied Earth Observations and Remote Sensing}, vol.~15, pp. 3365--3375, 2022.

\bibitem{FARAD}
\BIBentryALTinterwordspacing
{{FARADSAR public Release Data.}} [Online]. Available: \url{https://www.sandia.gov/radar/complex_data/FARAD_KA_BAND.zip.}
\BIBentrySTDinterwordspacing

\bibitem{gpa}
M.~Xu, H.~Wang, B.~Ni, Q.~Tian, and W.~Zhang, ``{Cross-domain detection via graph-induced prototype alignment},'' in \emph{Proceedings of the IEEE/CVF Conference on Computer Vision and Pattern Recognition}, 2020, pp. 12\,355--12\,364.

\bibitem{fie}
J.~Zhang, X.~Zhang, S.~Liu, B.~Pan, and Z.~Shi, ``{FIE-Net: Foreground Instance Enhancement Network for Domain Adaptation Object Detection in Remote Sensing Imagery},'' \emph{IEEE Transactions on Geoscience and Remote Sensing}, 2025.

\bibitem{mic}
L.~Hoyer, D.~Dai, H.~Wang, and L.~Van~Gool, ``{MIC: Masked image consistency for context-enhanced domain adaptation},'' in \emph{Proceedings of the IEEE/CVF conference on computer vision and pattern recognition}, 2023, pp. 11\,721--11\,732.

\bibitem{fcos}
Z.~Tian, C.~Shen, H.~Chen, and T.~He, ``{FCOS: Fully convolutional one-stage object detection},'' in \emph{Proceedings of the IEEE/CVF international conference on computer vision}, 2019, pp. 9627--9636.

\end{thebibliography}
    \bibliographystyle{IEEEtran}
\end{document}